\documentclass[10pt,twocolumn,letterpaper]{article}

\usepackage{cvpr}              % To produce the CAMERA-READY version

\usepackage{graphicx}
\usepackage{amsmath}
\usepackage{amssymb}
\usepackage{booktabs}
\usepackage{multirow}
\usepackage{array}
\usepackage{comment}
\usepackage{pifont}
\usepackage[pagebackref,breaklinks,colorlinks]{hyperref}

\usepackage[capitalize]{cleveref}
\crefname{section}{Sec.}{Secs.}
\Crefname{section}{Section}{Sections}
\Crefname{table}{Table}{Tables}
\crefname{table}{Tab.}{Tabs.}

\newcommand{\mysection}[1]{\vspace{3pt}\noindent\textbf{#1.}}

\begin{document}

\title{Learning Scene Flow  in 3D Point Clouds with Noisy Pseudo Labels  }

\author{Bing Li  \quad ChengZheng  \quad Guohao Li  \quad Bernard Ghanem\\
{King Abdullah University of Science and Technology}\\

}

\maketitle

%%%%%%%%% ABSTRACT
\begin{abstract}

We propose a novel scene flow method that captures 3D motions from point clouds without relying on ground-truth scene flow annotations. Due to the irregularity and sparsity of point clouds, it is expensive and time-consuming to acquire ground-truth scene flow annotations. Some state-of-the-art approaches train scene flow networks in a self-supervised learning manner  via approximating pseudo scene flow labels from point clouds.
However, these methods fail to achieve the performance level of fully supervised methods,  due to the limitations of point cloud such as sparsity and  lacking color information.
To provide an alternative, we propose a novel approach that utilizes monocular RGB images and point clouds to generate pseudo scene flow labels for training scene flow networks.
Our pseudo label generation module infers pseudo scene labels for point clouds by jointly leveraging rich appearance information in monocular images and geometric information of point clouds.
To further reduce the negative effect of noisy pseudo labels on the training,
we propose a noisy-label-aware training scheme by exploiting the geometric relations of points. Experiment results show that our method not only outperforms state-of-the-art self-supervised approaches but also outperforms some supervised approaches that use accurate ground-truth flows.
\end{abstract}

%%%%%%%%% BODY TEXT
\section{Introduction}
\label{sec:intro}

\begin{figure}[tb]
\centering
\includegraphics[bb=0 0 3214 1688, width=0.47\textwidth]{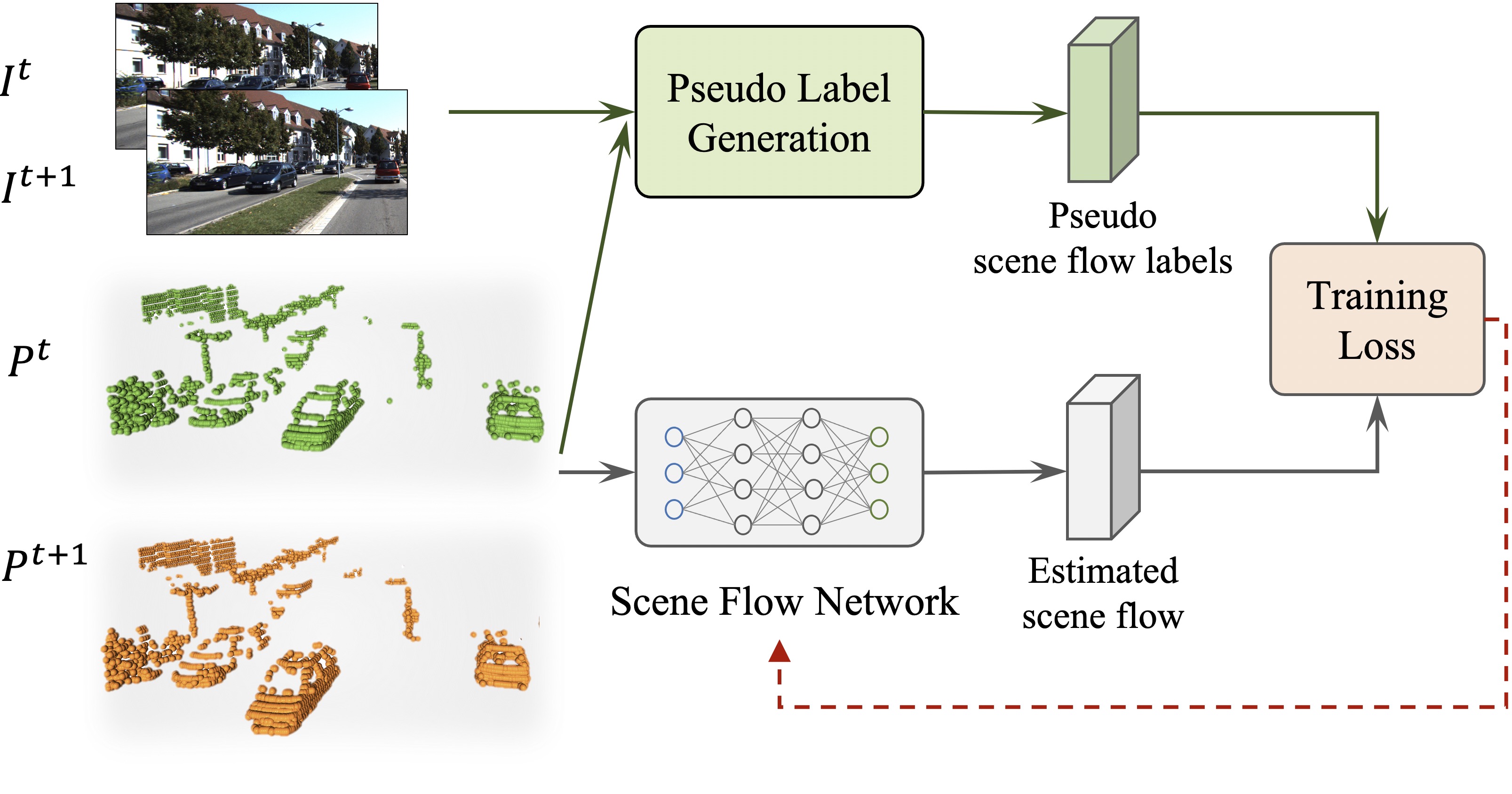}
\caption{\textbf{Illustration of our main idea}. We leverage multi-modality data (\ie monocular RGB images $I$ and point clouds $\mathcal{P}$) to generate pseudo scene flow labels for point clouds, such that   the reliance on ground-truth scene flow is  circumvented.  With the pseudo  labels,  our method trains a  scene flow network to estimate scene flow from point clouds.}
\vspace{-12pt}
\label{fig:fig1}	
\end{figure}

Scene flow estimation is to capture 3D motions of dynamic scenes, which is important for many applications such as robotics and autonomous driving. Recently, directly estimating scene flow from point clouds has received increasing attention. Nevertheless, it is challenging to estimate scene flow from point clouds, due to the sparsity and non-uniform density of point clouds.

Typical approaches  \cite{flot,pvraft,HPLFlowNet,FESTA} estimate scene flow from point clouds by training neural networks in a fully supervised manner, which relies on ground-truth scene flow annotation.
However, it is expensive and time-consuming to acquire ground-truth scene flows for real-world point clouds, since such annotations usually need to annotate 3D motions for every point of a point cloud. To alleviate this issue,  researchers resort to training the scene flow network on labeled synthetic data. 
However, these methods are limited in the effectiveness and generalization ability in real-world applications due to the domain gap between the synthetic data and real-world data.

Alternatively, some self-supervised methods  \cite{Mittal_2020_CVPR,pointpwc,FlowStep3DCVPR2021,randomwalk_sceneflow} train the scene flow network by constructing pseudo scene flow labels from point clouds. For example, Mittal \etal \cite{Mittal_2020_CVPR} approximated pseudo scene flow labels based on the coordinate differences of 3D points,  where the closest points to the next point cloud are treated as pseudo correspondence. These methods circumvent the reliance on ground-truth scene flows. However, they fail to achieve competitive performance compared with fully supervised approaches.

To achieve competitive performance without the need for ground-truth scene flow, we seek to generate high-quality pseudo scene flow labels for training scene flow networks. However, it is non-trivial to establish high-quality pseudo flow labels from the point cloud itself since a raw point cloud consists of only sparse point coordinates. \textit{Can we jointly leverage multi-modality data (\ie monocular RGB images and point clouds) to generate pseudo scene flow labels for point clouds}? (see Fig. \ref{fig:fig1}). 
Different from point clouds, monocular images contain rich information such as object appearance and detailed texture information. Such rich information provides discriminative cues for estimating 2D motions, which can be used to facilitate pseudo scene flow label generation.
The training dataset only needs to additionally provide RGB images captured by an affordable monocular camera, which is more accessible compared with expensive scene flow annotation.
Different from these methods \cite{Mittal_2020_CVPR,pointpwc,FlowStep3DCVPR2021,randomwalk_sceneflow}, our method leverages multi-modality data for generating pseudo scene flow labels.

However, it is challenging to generate pseudo scene flow labels.
We can not directly infer 3D motion from the 2D motion only relying on monocular images, despite the rich information of monocular images.
To address this issue, we propose a multi-modality-based pseudo scene flow generation module, through decomposing the 3D motion of a point into one 2D motion in the X-Y direction and another 1D motion in the Z direction.
We thereby can leverage monocular images to capture the 2D motions of points in the image plane and use point clouds to lift estimated 2D motions to pseudo scene flow labels (\ie. 3D motions). With the generated pseudo labels, our method can train scene flow networks on   point clouds without ground-truth scene flow. %annotation.

Another challenge is that our pseudo scene flow labels are inevitably noisy, compared with ground-truth ones, due to imperfect 2D motion estimation results. 
The noisy labels would negatively affect the training of the scene flow network, leading to the degradation of estimation accuracy.
To address this issue, we propose a noisy-label-aware learning scheme for scene flow estimation. Our scheme exploits the geometric information of point clouds to detect inaccurate labels in a soft manner. With the confidence scores of pseudo labels, we construct a training loss that less relies on pseudo labels with lower confidences when training scene flow networks.

The main contributions of our work are summarized as follows:
\textbf{(1)} We propose a novel method that trains the scene flow network without relying on ground-truth scene flow.
\textbf{(2)} We show how to leverage multi-modality data, \ie point clouds and monocular images, for generating pseudo scene flow labels.
\textbf{(3)} Our noisy-label-aware learning scheme effectively trains the scene flow network by reducing the negative effect of the inherent noise of pseudo labels during training.
\textbf{(4)} Experiment results show that our method not only outperforms state-of-the-art self-supervised approaches but also outperforms some supervised approaches that use accurate ground-truth annotations, even though our pseudo labels are noisy and inferior.
\textbf{(5)} The pseudo scene flow labels allow us to train the scene flow network on large-scale real-world LiDAR data without ground-truth scene flow, reducing the domain gap.

\begin{figure*}[tb]
\centering
\includegraphics[bb=0 0 6612 1928, width=0.95\textwidth]{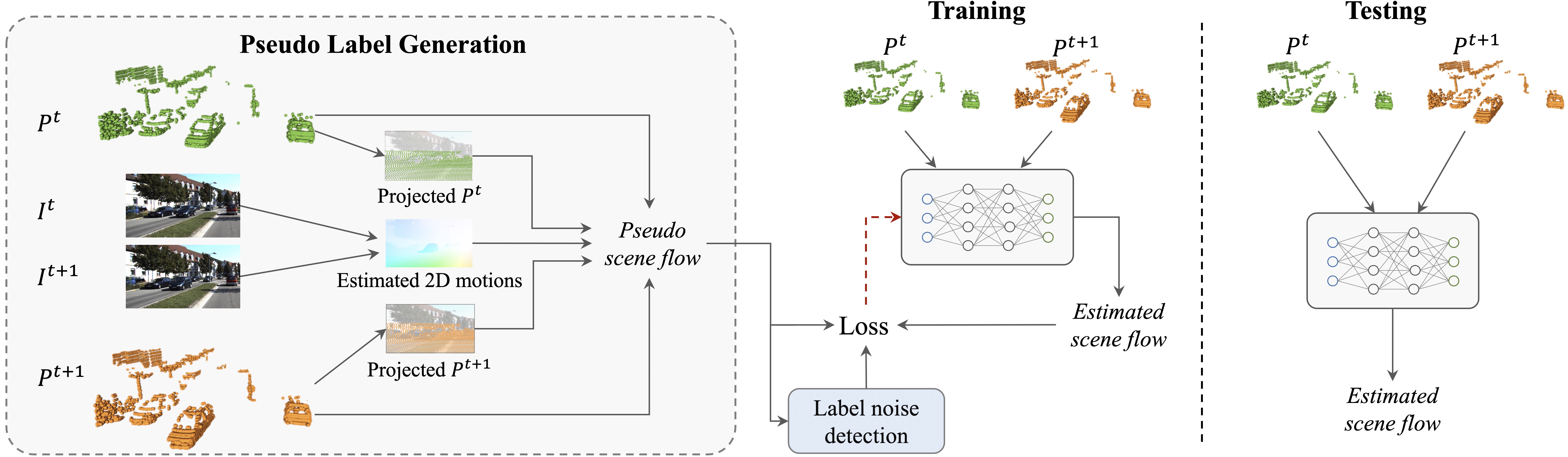}
\caption{\textbf{The framework of our scene flow estimation method.} Our method leverages two-modality data, \ie monocular images $I$ and point clouds $\mathcal{P}$, to generate pseudo scene flow labels for point clouds. With the pseudo labels, we train a scene flow network for estimating scene flow from point clouds, while using the proposed noise detection module to reduce the negative effect of noisy labels. During the inference, our scene flow network takes only point clouds as input and outputs estimated scene flow.}
\label{framework}
\vspace{-13pt}	
\end{figure*}

\section{Related Work}

\mysection{Optical Flow}
Optical flow estimation is to capture 2D motions from monocular video frames.  As a fundamental tool for 2D scene understanding, many works have extensively studied optical flow estimation. Early approaches \cite{horn1981determining,black1993framework,zach2007duality,weinzaepfel2013deepflow,brox2009large,ranftl2014non} estimate optical flow via energy minimization. Inspired by the success of  neural networks, many data-driven approaches   \cite{dosovitskiy2015flownet,mayer2016large,ilg2017flownet,hui2018liteflownet,hui2020liteflownet3,PWCNet,teed2020raft} are proposed to estimate optical flow by training neural networks.

\mysection{Scene Flow from Stereo and RGB-D Videos}
Many methods \cite{chen2020consistency,vogel2013piecewise,wedel2008efficient,ilg2018occlusions,jiang2019sense,teed2020raft-3d,huguet2007variational} are proposed to estimate scene flow from stereo videos or RGB-D frames. These works formulate scene flow estimation as a problem jointly estimating disparity map and optical flow from stereo frames \cite{MIFDB16}. Recent works \cite{ilg2018occlusions,ma2019drisf,chen2020consistency} propose neural-network-based models to estimate scene flow from stereo video. Some works \cite{quiroga2014dense,sun2015layered} focus on estimating scene flow from RGB-D videos. Methods \cite{Hur:2021:SSM,Liu:2019:unrigid,Hur:2020:SSM} design self-supervised loss such as photometric consistency loss to train networks in a self-supervised manner.

\mysection{Supervised Scene Flow From Point Clouds}  
Estimating scene flow for point cloud has attracted increasing attention. Most works \cite{flownet3d,pvraft,FESTA,flot,HPLFlowNet,Li2021CVPR12} address the problem of point-cloud-based scene flow estimation through fully supervised learning.
Different from traditional methods \cite{dewan2016rigid,ushani2017learning}, 
FlowNet3D~\cite{flownet3d} proposed a network to learn 3D scene flow from point clouds in an end-to-end manner, where a flow embedding layer was introduced.  
Based on FlowNet3D~\cite{flownet3d}, FlowNet3D++~\cite{FlowNet3Dplus} incorporated geometric constraints to improve FlowNet3D. HPLFlowNet~\cite{HPLFlowNet} introduced Bilateral Convolutional Layers for estimating scene flow by projecting point clouds into permutohedral lattices.  
FLOT \cite{flot} treated scene flow estimation as a correspondence matching problem, and employ  optimal transport to find correspondences between the point clouds. FESTA \cite{FESTA} addressed the issue of farthest point sampling for FlowNet3D~\cite{flownet3d} by introducing spatial attention layers and temporal attention layers.
Li \etal \cite{Li2021CVPR12} enforced point-level and
region-level consistency by deploying CRFs based relation modules.
Gojcic~\etal~\cite{weaklyrigidflow} proposed a weakly-supervised method for scene flow estimation, by using ground-truth ego motions and foreground and background annotations. 
Different from these methods, we focus on learning scene flow without ground-truth scene flow.

\mysection{Self-Supervised Scene Flow From Point Clouds}
Some works~\cite{pointpwc,FlowStep3DCVPR2021,Mittal_2020_CVPR} have explored unsupervised or self-supervised learning for estimating scene flow from point clouds. To train the scene flow network without ground-truth annotation, Mittal \etal 
\cite{Mittal_2020_CVPR} approximated pseudo scene flow labels based on the coordinate differences.
Similarly, Wu \etal \cite{Mittal_2020_CVPR} chose the Chamfer loss as the proxy loss of  self-supervised learning. Besides Chamfer loss, Wu \etal \cite{Mittal_2020_CVPR} used smoothness constraint and laplacian regularization to enforce the local consistency of predicted scene flow.
In addition, PointPWC \cite{pointpwc} proposed a new scene flow network by introducing cost volume, upsampling and warping modules, inspired by optical flow method PWC-Net~\cite{PWCNet}.
FlowStep3D\cite{FlowStep3DCVPR2021} took the inspiration of RAFT \cite{teed2020raft} and recurrent architecture to iteratively refine estimate scene flow.
Besides point coordinates, SelfPF \cite{randomwalk_sceneflow} adopted additional measures such as surface normal of point cloud to generate pseudo labels, where optimal transport and random walk were employed to find pseudo correspondences.

Different from all these self-supervised methods, our method explores how to generate pseudo scene flow labels from multi-modality data, \ie monocular RGB images and point clouds. This is also inspired by multi-sensor-based methods on object tracking \cite{mmMOT2019ICCV} , object detection \cite{qi2020imvotenet,meyer2019sensor} and point cloud segmentation\cite{FuseSeg,Zhuang_2021_ICCV}. In addition, we detect noisy labels and propose a noisy-label-aware training scheme to reduce the negative effect of noisy labels on the training, different from existing methods. 

%-------------------------------------------------------------------------

\section{Method}

\begin{figure*}[!tp]
\centering	
\centering	
\begin{minipage}[t]{0.19\textwidth}
\centering
\includegraphics[bb=0 0 988  996, width=1\textwidth]{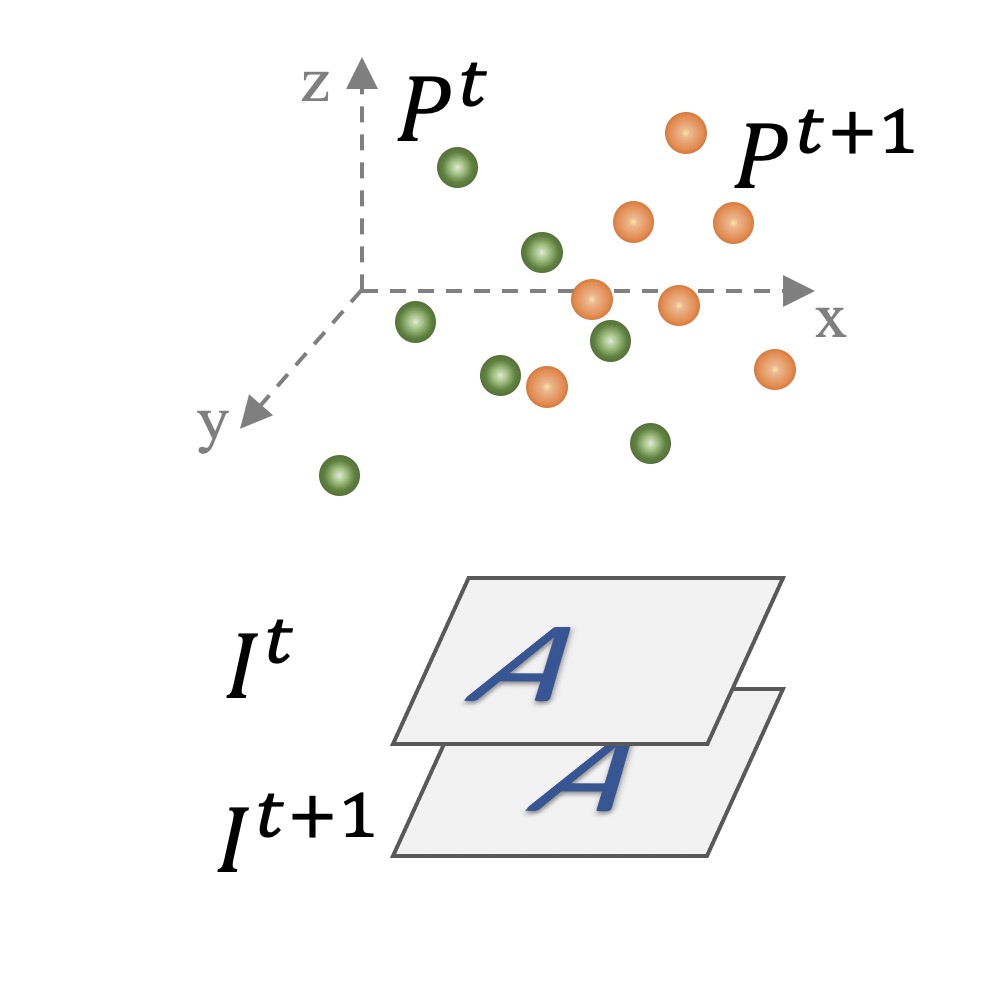}			
\scriptsize (a)
\end{minipage}
\begin{minipage}[t]{0.19\textwidth}
\centering
\includegraphics[bb=0 0 988  996, width=1\textwidth]{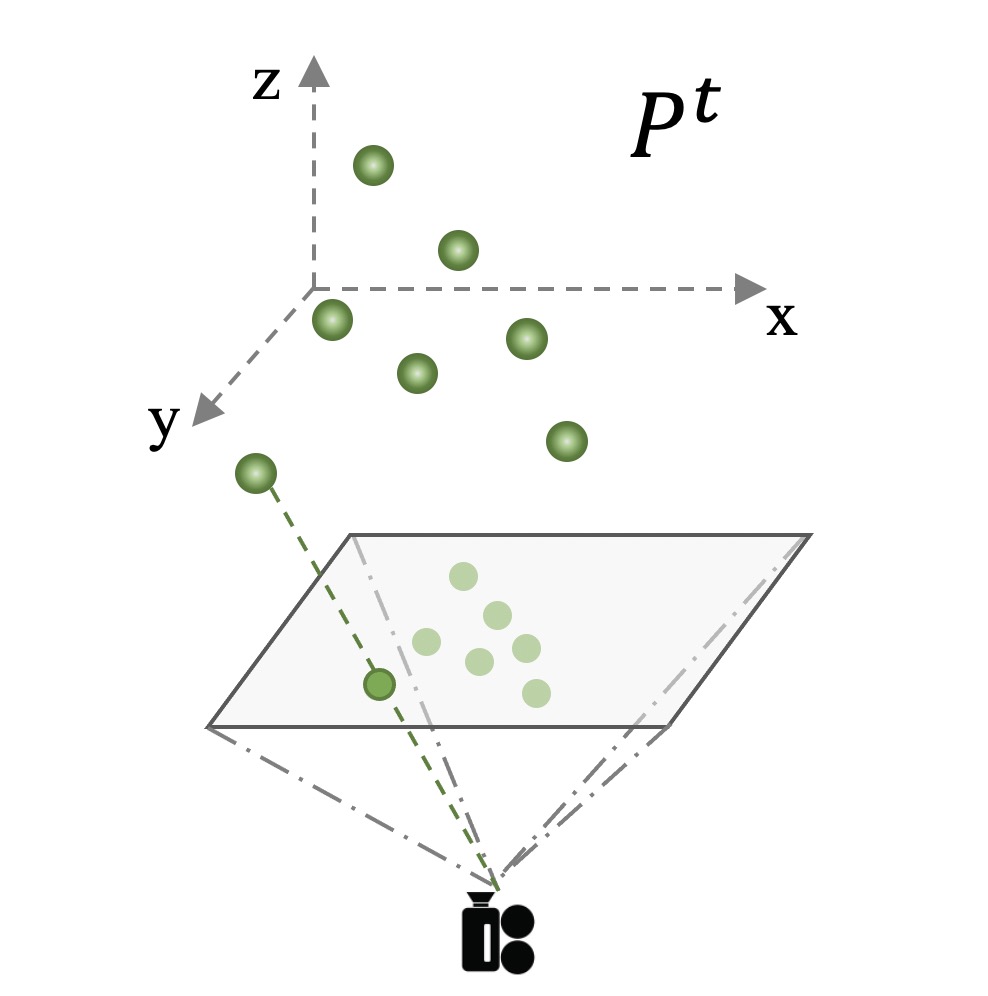}			
\scriptsize (b)
\end{minipage}
\begin{minipage}[t]{0.19\textwidth}
\centering
\includegraphics[bb=0 0 988  996, width=1\textwidth]{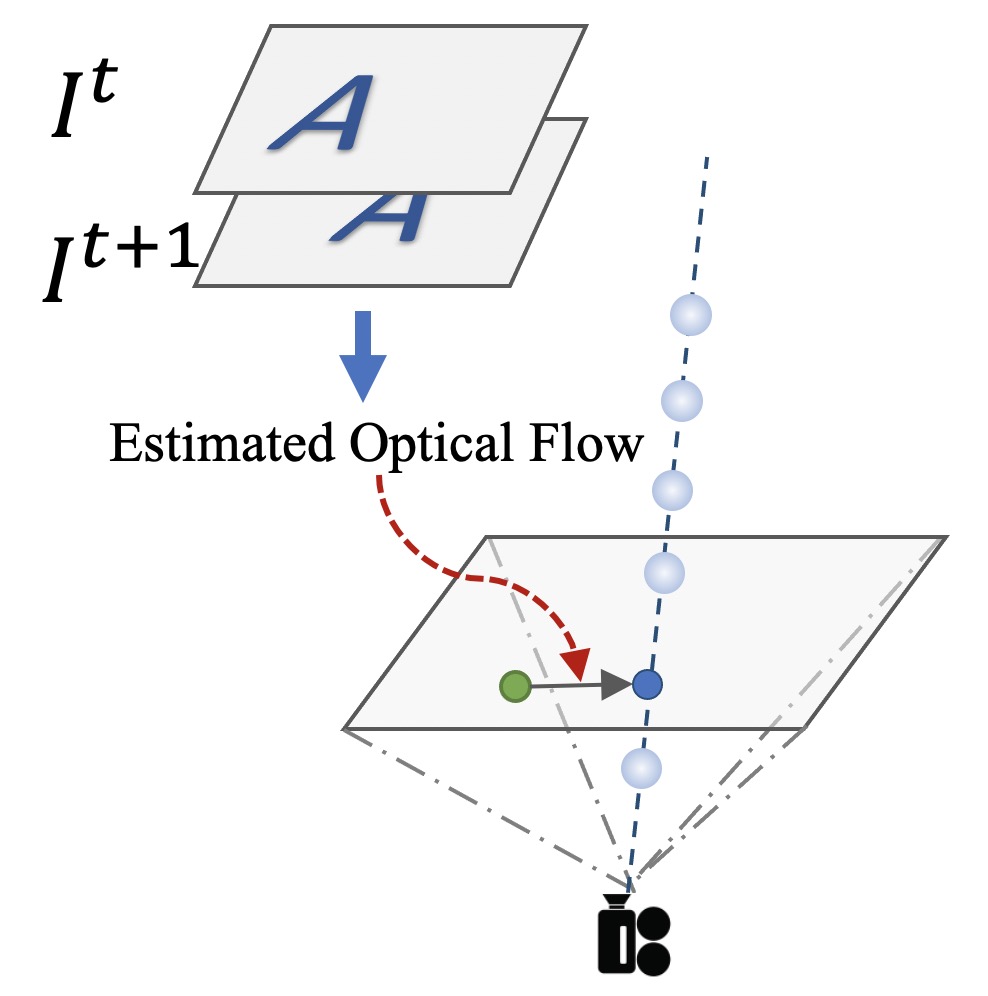}			
\scriptsize (c)
\end{minipage}
\begin{minipage}[t]{0.19\textwidth}
\centering
\includegraphics[bb=0 0 988  996, width=1\textwidth]{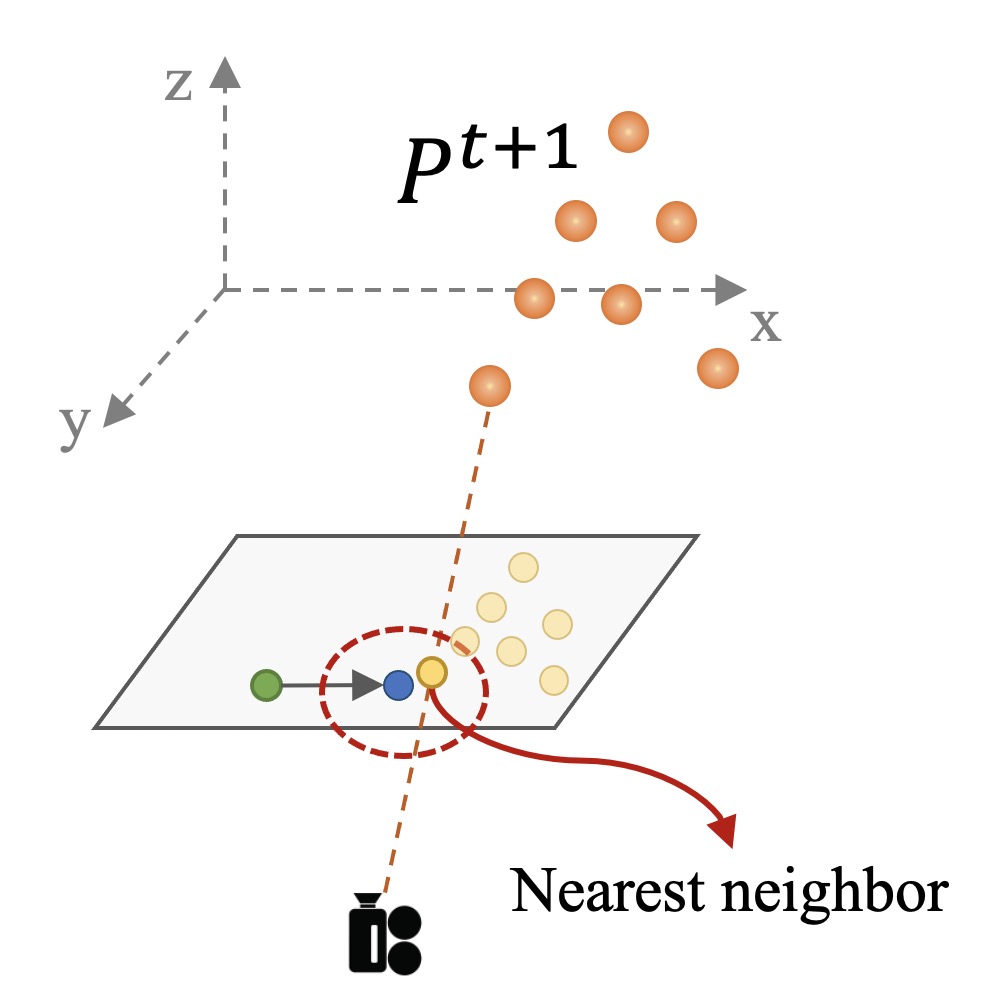}			
\scriptsize (d)
\end{minipage}
\begin{minipage}[t]{0.19\textwidth}
\centering
\includegraphics[bb=0 0 988  996, width=1\textwidth]{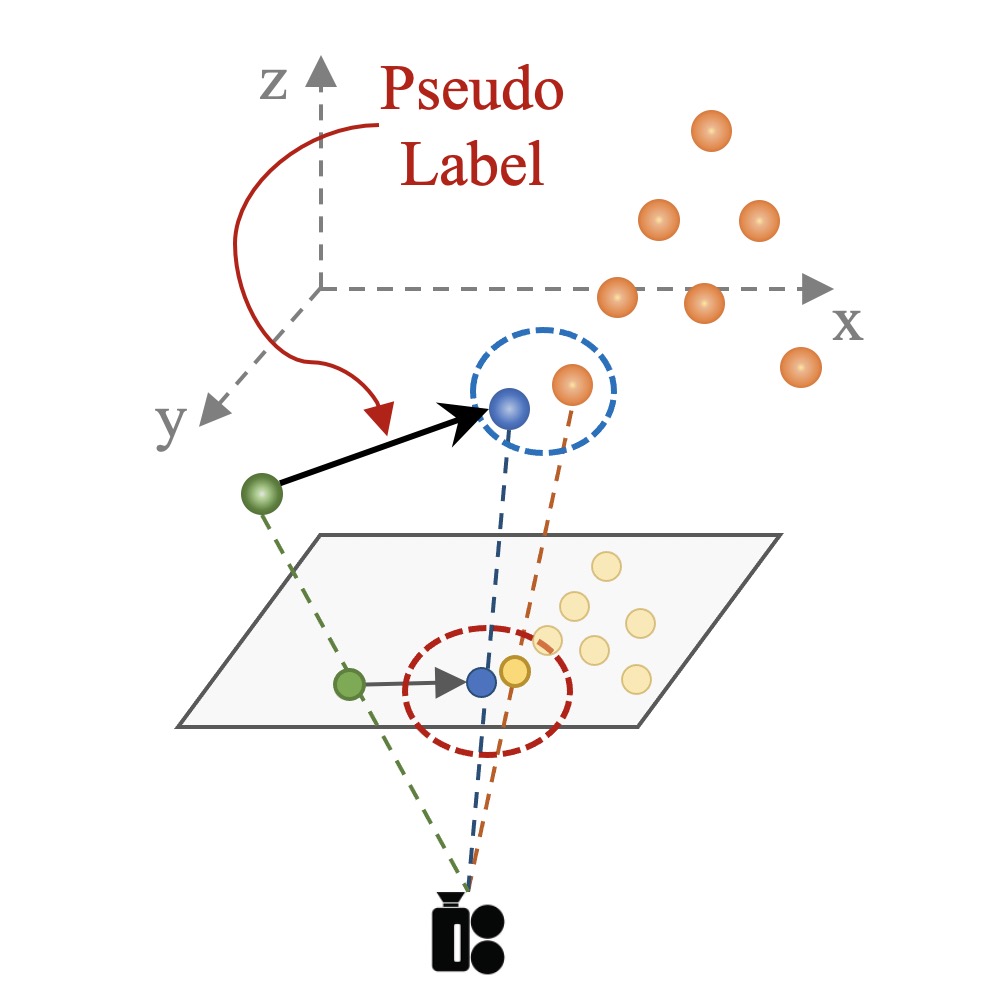}			
\scriptsize (e)
\end{minipage}
\caption{\textbf{Illustration of pseudo scene flow label generation.} (a) The input point clouds and monocular images, where point clouds $\mathcal{P}^t$ are colored by green and $\mathcal{P}^{t+1}$ by orange. (b) $\mathcal{P}^t$ is projected to the image plane via a perceptive projection. (c) Given projected point $\bar{p}_i^t$ of $\mathbf{p}^t_i$, we infer its 2D correspondence using estimated optical flows from images $I^t$ and $I^{t+1}$. Yet, according to  the 2D correspondence, there are many  candidate 3D correspondences (indicated by light blue balls). (d) We search for a point $\mathbf{p}^{t+1}$ in $\mathcal{P}^{t+1}$ whose projected point is nearest to the 2D correspondence. (e) With  $\mathbf{p}^{t+1}$, we obtain the 3D correspondence and generate the pseudo labels.
}
\label{fig: Pseudo label generation}
\vspace*{-12pt}%-12
\end{figure*}

Different from existing supervised scene flow estimation approaches, we propose a method that learns scene flow from point clouds without using ground-truth scene flow (see Fig. \ref{framework}). 
To circumvent the reliance on ground truth scene flow, we propose a pseudo label generation module that infers pseudo labels from monocular images and point clouds by leveraging the complementary information of these multi-modality data (see Fig. \ref{fig: Pseudo label generation}).
With generated scene flow labels, our method trains scene flow networks on only point clouds.
Nevertheless, the generated pseudo labels are still noisy, compared with ground-truth ones. To address this issue, 
we design training losses to reduce the negative effect of noisy labels on the training in Sec. \ref{sec.noisy learning}.

%%%%%%%%%%%%%%%%%%%%%%%%%%%%%%%%%%%%%%%%%%%%%%
%%%%%%%%%%%%%%%%%%%%%%%%%%%%%%%%%%%%%%%%%%%%%%
\subsection{Pseudo Label Generation via Multi-modalities}
We aim to generate pseudo scene flow labels for point clouds, such that scene flow networks can be trained without ground-truth ones. We argue that it is non-trivial to establish high-quality pseudo flow labels from point clouds themselves. Instead, we propose a Multi-modalities based Pseudo Label Generation (M-PLG) module that generates pseudo scene flows for point clouds with the assistance of monocular RGB images.

\mysection{Notations} The pseudo label generation module takes two point clouds and their 
synchronized monocular images as input.
Let  $\{     \left\langle\mathcal{P}^t,{I}^t\right\rangle, \left\langle\mathcal{P}^{t+1} ,{I}^{t+1}\right\rangle\}$ denotes an input sample captured from a dynamic scene, where $I^t$ is a monocular RGB image captured by a camera at time $t$, and $\mathcal{P}^t$ is a point cloud at $t$.  The point cloud $\mathcal{P}^t=\{\mathbf{p}^t_i\}_{i=1}^{N}$ consists of $N$ points, where the $i$-th point $\mathbf{p}^t_i\in \mathbb{R}^3$ is represented by its XYZ coordinates. The M-PLG module outputs a pseudo flow label $\hat{\mathbf{f}}^t_{i}\in\mathbb{R}^3$ for each point $\mathbf{p}^t_i$ in $\mathcal{P}^t$.

\mysection{Problem Definition of Pseudo Label Generation} 
Pseudo scene flow labels are to describe 3D displacements for each point from point cloud $\mathcal{P}^t$ to $\mathcal{P}^{t+1}$.
Given a point $\mathbf{p}^t_i \in \mathcal{P}^t$, we construct its pseudo scene flow label by inferring the position of its pseudo 3D correspondence\footnote{The location of $\mathbf{p}^t_i$'s correspondence often does not coincide any points in $\mathcal{P}^{t+1}$, since point cloud is sparse, different from images.} in the point cloud  $\mathcal{P}^{t+1}$. With the position of the correspondence, pseudo labels are calculated as follows:
\begin{equation}
\hat{\mathbf{f}}_i^t=\hat{\mathbf{c}}^{t+1}_i-\mathbf{p}^t_i
\label{eq:pseudo_label}
\vspace*{-2pt}%-12
\end{equation}
where $\hat{\mathbf{c}}^{t+1}_i$ is the pseudo correspondence of $\mathbf{p}^t_i$ in $\mathcal{P}^{t+1}$.

We address the problem of inferring the correspondence $\hat{\mathbf{c}}^{t+1}_i$ by additionally leveraging monocular images $I^t$ and $I^{t+1}$, since $I^t$ and $\mathcal{P}^t$ are captured from the same scene. To this end, we decompose the 3D motion of point $\mathbf{p}^t_i$ into one 2D motion in the X-Y direction and another 1D motion in the Z direction (\ie depth). 
For X-Y direction, we leverage images $I^t$ and $I^{t+1}$ to estimate the 2D correspondences (\ie 2D motions) of point cloud $\mathcal{P}^t$ in the image plane at $t+1$, through projecting $\mathcal{P}^{t}$ to the image plane (see Fig. \ref{fig: Pseudo label generation}{\color{red}b} and \ref{fig: Pseudo label generation}{\color{red}c}). 
Compared with point clouds, images contain rich information such as texture and color, which is discriminative and helpful for finding 2D correspondences.
However, according to such 2D correspondence, there are many candidate 3D correspondences in $\mathcal{P}^{t+1}$, since the depth ($Z$ value) of 2D correspondence is unknown. 
To address this issue, we propose to exploit the relations of 2D correspondences and point cloud $\mathcal{P}^{t+1}$  to infer the 3D correspondences.
Therefore, we propose M-PLG module, whose overall pipeline is illustrated in Fig. \ref{fig: Pseudo label generation}.
Below we discuss each step of this module in detail.

\mysection{Projecting Point Cloud via Perspective Projection} Given a point $\mathbf{p}^t_i$ in point cloud $\mathcal{P}^t$, we project it to the image plane, such that monocular images can be leveraged for estimating the 2D motion of $\mathbf{p}^t_i$.
In particular, we first transform $\mathbf{p}^t_i$ into  the camera coordinate system of image $I^t$ 
through perspective projection. Let $\mathbf{p}^t_i=[X,Y,Z]$ denotes the position of  $\mathbf{p}^t_i$,
the projected point of $\mathbf{p}^t_i$ to the camera coordinate system is calculated as follows:

\begin{equation}
[x,y,z]^T= \mathbf{M} \cdot [\mathbf{p}^t_i,1]^T
\end{equation}
where $\mathbf{M} \in \mathbb{R}^{3\times4}$ is the perspective projection matrix that consists of a rotation matrix and a translation vector. For  multi-sensor data (\ie, point clouds captured by a Lidar sensor and RGB images by a camera sensor), we obtain the perspective projection matrix through camera and Lidar sensor calibration \cite{geiger2012automatic}.

We then map the projected point  $[x,y,z]$ to the image coordinate system \ie $\bar{p}^{t}_i=(x/z,y/z)$, to
obtain 2D projected point $\bar{p}^{t}_i$ of  $\mathbf{p}^t_i$ in the image plane.

\mysection{Estimating 2D Correspondences} For a 2D projected point $\bar{p}^{t}_i$, we predict its 2D correspondence in the image plane using monocular images $I^t$ and $I^{t+1}$. We employ optical flow algorithms (\eg \cite{teed2020raft}) to estimate 2D motions from $I^t$ to $I^{t+1}$. According to the position of  projected point $\bar{p}^{t}_i$, we obtain its 2D optical flow ${f}^{I}_i$ and further estimate its 2D correspondence $\bar{c}^{t+1}_i$ in the image plane as 
$\bar{c}^{t+1}_i=\bar{p}^{t}_i + {f}^{I}_i  $.

Note that point $\mathbf{p}^t_i$ has many candidate 3D correspondences at $t+1$, according to 2D correspondence $ \bar{c}^{t+1}_i$, since the depth is unknown. 
We can not infer 3D correspondence of a point from its 2D correspondence, if we only rely on the information of monocular images.

\mysection{Inferring 3D correspondences} Without the depth (\ie $Z$) value, we cannot lift the 2D correspondences to 3D. Instead, we exploit point cloud $\mathcal{P}^{t+1}$ for inferring the $Z$ value of the 3D correspondence based on two observations. First,
we observe that the true 3D correspondence of $\mathbf{p}^{t}_i\in \mathcal{P}^t$ either coincides or is very close to a point in $\mathcal{P}^{t+1}$. Second, the nearest neighboring points often have similar $Z$ values in the 3D space.
That is, if the true 3D correspondence of $\mathbf{p}^{t}_i$ is very close to a point  $\mathbf{p}^{t+1}_j$ in $\mathcal{P}^{t+1}$, we can approximately use the $Z$ value of $\mathbf{p}^{t+1}_j$ as that of the true 3D correspondence. 
However, the true 3D correspondence is unknown. Since 2D correspondences are known,  we search the nearest point in the image plane to infer the $Z$  value. In particular, we  project $\mathcal{P}^{t+1}$ to the image plane via a perspective projection. Then, for $\mathbf{p}^{t}_i$, we search for a 2D projected point of $\mathcal{P}^{t+1}$ which is the nearest  to its 2D correspondence $\bar{c}^{t+1}_i$  as:
\begin{equation}
{\bar{p}^{t+1}_k} =\arg \min_{\bar{p}^{t+1}_j\in \bar{\mathcal P}^{t+1}} ||\bar{p}^{t+1}_j -\bar{c}^{t+1}_i||_2
\vspace*{-3pt}%-12
\end{equation}
where $\bar{p}^{t+1}_k$ is the 2D projected point of $\mathbf{p}^{t+1}_k$, and $\bar{\mathcal P}^{t+1}$ is the set including all 2D projected points of $\mathcal{P}^{t+1}$.   

By using the $Z$ value of $\mathbf{p}^{t+1}_k$ to approximate that of 3D correspondence $\hat{\mathbf{c}}^{t+1}_i$, we transform the 2D correspondence $\bar{c}^{t+1}_i$ to a 3D correspondence $\hat{\mathbf{c}}^{t+1}_i$ (the details are given in the supplementary). We then infer the pseudo scene flow label of $\mathbf{p}^{t}_i$ by substituting $\mathbf{c}^{t+1}_i$ in Eq. \ref{eq:pseudo_label}.

Note that the generated pseudo labels are inevitably  noisy. For example, the optical flow estimated results are imperfect, leading to inaccurate pseudo labels. 
To address this issue, we propose a noisy-label-aware learning scheme.

%%%================================================================================
\subsection{Learning Scene Flow with Noisy Pseudo Labels}
\label{sec.noisy learning}
We propose a noisy-label-aware learning scheme to train scene flow networks on point clouds with noisy pseudo scene flow labels, different from fully supervised methods.

Our generated pseudo scene flow labels are noisy, which would negatively degrade the training of scene flow networks. To address this issue, we aim to decrease the influence of inaccurate pseudo labels during training. This is different from supervised or self-supervised learning of existing works treating all scene flow labels equally.
To achieve this aim, we leverage the information of point clouds to detect  inaccurate pseudo labels that wrongly describe the 3D motions of the point, and assign low confidence scores to these labels. We then use the confidence scores of pseudo labels to design our training loss.

\mysection{Label Noise Detection}
Our label noise detection module detects noisy labels in a soft manner, which predicts a confidence score for each   pseudo label.
We exploit the information of point clouds to estimate the reliability of each pseudo label $\hat{\mathbf{f}}_i$. In particular, the true 3D correspondence of $\mathbf{p}^{t}_i \in \mathcal{P}^t$ either coincides or is close to a point in $\mathcal{P}^{t+1}$. Based on this observation,  
we detect inaccurate pseudo label according to the nearest point in $\mathcal{P}^{t+1}$.  
In other words, if a pseudo 3D correspondence $\hat{\mathbf{c}}^{t+1}_i=\hat{\mathbf{f}}_i + \mathbf{p}^{t}_i $ (calculated by pseudo label $\hat{\mathbf{f}}_i$), is far away from its nearest point in $\mathcal{P}^{t+1}$, 
pseudo label $\hat{\mathbf{f}}_i$ is probably to be inaccurate. 
To reduce computational cost, we search for the nearest point in the 2D projected point of $\mathcal{P}^{t+1}$ in the image plane.
Therefore, given a pseudo label $\hat{\mathbf{f}}_i$, we define its initial confidence score $w_i$ according to the distance of pseudo 3D correspondence and its nearest point $\mathbf{p}^{t+1}_j\in\mathcal{P}^{t+1}$ in the image space:
\begin{equation}
w_i=f(d_i) =f(||\bar{p}^{t+1}_j -\bar{c}^{t}_i||_2),
\end{equation}
where $\bar{c}^{t}_i$ is the 2D projected point of  $\hat{\mathbf{c}}^{t+1}_i$ in the image plane, and  $\bar{p}^{t+1}_j$ is a 2D projected point of $\mathbf{p}^{t+1}_j\in\mathcal{P}^{t+1}$ nearest to $\bar{c}^{t}_i$.
In this paper, we simply set $f(\cdot)$ to be:
\begin{equation}
f(d)   =
\begin{cases}
&1, \quad if \quad d <\theta\\
& \frac{1}{d}, \quad otherwise
\end{cases}
\vspace*{-3pt}%-12
\end{equation}
If the distance $d$ of $\bar{p}^{t+1}_j $ and $\bar{c}^{t}_i$ is larger,  $\hat{\mathbf{f}}_i$ is of higher probability to be inaccurate. We thereby assign a lower value  to the confidence score of  $\hat{\mathbf{f}}_i$.  We set $\theta=2$ in our experiments.

We further refine the confidence scores for pseudo labels, since we aim to reduce the negative effect of inaccurate pseudo labels as much as possible.
Otherwise, pseudo labels with large errors would make the training unstable, leading to poor performance. We refine the confidence scores by using local geometric information of point cloud.
In particular, given a point $\mathbf{p}^{t}_i $, if the pseudo labels of its neighbor are inaccurate and its pseudo label is similar to those labels, we argue that its pseudo label is possibly to be inaccurate. Therefore, given a pseudo label $\hat{\mathbf{f}}_i$, we update its confidence score according to the pseudo labels and  confidence scores of  $\mathbf{p}^{t}_i$'s neighbors: 
\begin{equation}
w_i^{u}  = \lambda w_i+ (1-\lambda )\frac{1}{K} \sum_{k} w_k \cdot a_{ik}
\vspace*{-3pt}%-12
\end{equation}
where $ a_{ik}=\exp(-\frac{||\hat{\mathbf{f}}_k- \hat{\mathbf{f}}_i||_2}{\tau})$ measuring the pseudo label similarity of between $\hat{\mathbf{f}}_i$ and $\hat{\mathbf{f}}_k$, $\tau$ is a predefined hyper-parameter, $\lambda \in [0,1]$ is a weighted parameter that controls the strength of the updating progress. 
Thus, the confidence score of $\hat{\mathbf{f}}_i$ would be decreased, if $\hat{\mathbf{f}}_i$ is similar to the pseudo labels of neighbors and these labels are inaccurate.

\mysection{Weighted training loss} We build the weighted training loss using the confidence scores of pseudo labels as the weights.
The weighted loss is formulated as follows:

\begin{equation}
L =\sum_{i=1} w_i^u||\hat{\mathbf{f}}_i-{\mathbf{f}}_i||_1
\label{eq:training}
\vspace*{-3pt}%-12
\end{equation}
where ${\mathbf{f}}_i$ is the predicted scene flow of point $\mathbf{p}^t_{i}$ estimated from scene flow networks. By using confidence scores $w^u$ as loss weights, pseudo labels with lower confidences play less important role in training. The weight of some pseudo labels may be too small, leading to the absence of supervision. We can address this by  constraining a point and   its neighbors to have similar predicted scene flows.
Since our method provides pseudo scene flow labels, we can directly employ the scene flow networks of existing point-cloud-based self-supervised or supervised methods \eg, \cite{pointpwc,pvraft,flot,FlowStep3DCVPR2021}.

\begin{table*}[t]
\centering
%\begin{tabular}{|c|c|c|}
\small
\begin{tabular}{p{1.3cm}| p{2.6 cm}  |p{0.8cm}<{\centering}|  p{1.9cm}<{\centering} p{1.9cm}<{\centering} p{1.9cm}<{\centering}p{1.9cm}<{\centering} }
\toprule
Dataset & Method & Sup. & EPE3D $\downarrow$ & Acc3DS $\uparrow$ & Acc3DR $\uparrow$ & Outliers $\downarrow$ \\
\midrule
\multirow{5}{*}{FT3D} 
& FlowNet3D \cite{flownet3d}       & \textit{Full} &   0.114 &   0.412 &   0.771 &   0.602 \\
& HPLFlowNet \cite{HPLFlowNet}     &\textit{Full} &   0.080 &   0.614 &   0.855 &   0.429 \\
& PointPWC \cite{pointpwc}          & \textit{Full}&   0.059 &   0.738 &   0.928 &   0.342 \\
& EgoFlow \cite{tishchenko2020self} & \textit{Full}&   0.069 &   0.670 &   0.879 &   0.404 \\
& FLOT  \cite{flot}                 & \textit{Full}&   0.052 &   0.732 &   0.927 &   0.357 \\
& PV-RAFT\cite{pvraft}             & \textit{Full}      &  0.046 & 0.817& 0.957& 0.292 \\
& Rigid3DSF\cite{weaklyrigidflow} & \textit{Full} & 0.052 & 0.746& 0.936& 0.361\\
\cline{2-7}
& PointPWC \cite{pointpwc}          & \textit{Self}&   0.121 & 0.324 & 0.674 &0.688 \\
& SelfPF \cite{randomwalk_sceneflow} & \textit{Self} &0.112	&  0.528	&  0.794	& 0.409 \\
& FlowStep3D \cite{FlowStep3DCVPR2021} & \textit{Self} &0.085 & 0.536 & 0.826 &0.420 \\
\cline{2-7}
& \textbf{Ours}       & \textit{Pseudo}&   \textbf{0.068} &  \textbf{0.628} &   \textbf{0.881} &   \textbf{0.438} \\

\midrule
\multirow{5}{*}{S-KITTI} 
& FlowNet3D \cite{flownet3d}     & \textit{Full}    &   0.177 &   0.374 &   0.668 &   0.527 \\
& HPLFlowNet \cite{HPLFlowNet}   & \textit{Full}    &   0.117 &   0.478 &   0.778 &   0.410 \\
& PointPWC \cite{pointpwc}       & \textit{Full}    &   0.069 &   0.728 &   0.888 &   0.265 \\
& EgoFlow \cite{tishchenko2020self}  & \textit{Full}&   0.103 &   0.488 &   0.822 &   0.394 \\
& FLOT \cite{flot}              & \textit{Full}      &   0.056 &   0.755 &   0.908 &   0.242 \\
& PV-RAFT\cite{pvraft}             & \textit{Full}      &  0.056 & 0.823& 0.937& 0.216 \\
& Rigid3DSF\cite{weaklyrigidflow} & \textit{Full} & 0.042 & 0.849 & 0.959 &0.208\\
\cline{2-7}
& PointPWC \cite{pointpwc}          & \textit{Self}&  0.255 & 0.238& 0.496& 0.686 \\
& SelfPF \cite{randomwalk_sceneflow} & \textit{Self} &0.112& 0.528& 0.794& 0.409 \\
& FlowStep3D \cite{FlowStep3DCVPR2021} & \textit{Self} &0.102 & 0.708 & 0.839 &0.246 \\
\cline{2-7}
& \textbf{Ours}            & \textit{Pseudo}&  \bf0.058 &\bf0.744 & \bf0.898 &\bf	0.246	 \\
\bottomrule
\end{tabular}
\caption{\textbf{Comparison results  on FT3D and S-KITTI datasets.} For FT3D, \textit{Full} refers to fully-supervised training under the guidance of ground-truth scene flow, while \textit{self} means that training without such labels. 
As to  S-KITTI, both \textit{Self} and\textit{Full}  means  the model trained  on  FT3D  is directly evaluated   on S-KITTI.  
Without ground-truth  flow labels, our method not only achieves the highest performances in all metrics among self-supervised methods, but also outperforms some supervised methods. }
\label{tab:comparison}
\vspace*{-11pt}
\end{table*}
\section{Experiments}

\mysection{Dataset} We conduct experiments on FlyingThings3D~\cite{flythings3d} and StereoKITTI datasets that are widely used in recent methods \eg \cite{flot,pointpwc,HPLFlowNet,pvraft,FlowStep3DCVPR2021} for a fair comparison. In addition, we evaluate our method on a real-world LiDAR dataset.

\textit{FT3D (FlyingThings3D)}~\cite{flythings3d} is a large-scale synthetic stereo video dataset. The stereo videos are generated by assigning various motions to synthetic objects, where objects are from ShapeNet \cite{shapenet}. Following the preprocessing in \cite{HPLFlowNet,pointpwc}, 
we use camera parameters and ground-truth annotations to generate point clouds and ground truth scene flow. 
For a fair comparison, we randomly sample 8192 points per point cloud and remove points whose depth is larger than 35m, like existing methods \cite{FlowStep3DCVPR2021,flot}.

\textit{S-KITTI (StereoKITTI)}~\cite{Menze2018JPRS,minkowskiSPC} is a real-world scene flow dataset generated from stereo cameras.   
Following the data preprocessing of HPLFlowNet \cite{HPLFlowNet}, we obtain 142 pairs of point clouds using ground-truth disparity maps and optical flows. All these pairs are used for testing, where 8192 points are randomly sampled per point cloud. 
Following the setting of existing methods \cite{flot,pointpwc,HPLFlowNet}. 
we remove ground points according to point heights ($<-1.4$m).

\textit{L-KITTI (LidarKITTI)} is real-world LiDAR dataset constructed from the raw data of KITTI \cite{geiger2013vision}. \textit{L-KITTI} consists of point cloud sequences and image sequences captured from multiple scenes, where point clouds are captured by Velodyne 3D laser scanner. 
Following \cite{weaklyrigidflow}, we obtain the ground-truth scene flow for only 142 pairs of point clouds, since they correspond to the same scene of \textit{S-KITTI}.  
We use these 142 pairs of point clouds for testing.  The remaining 8201 pairs of points clouds are used for training, where testing point clouds and their temporally adjacent ones are excluded. 
Similar to S-KITTI, we remove ground points.
The visual patterns of LiDAR point clouds in L-KITTI  are rather dissimilar to that in S-KITTI and FT3D, which is challenging for scene flow estimation methods.

\mysection{Evaluation Metrics} To evaluate the performance of our approach, we use four standard evaluation metrics widely used in point-cloud-based scene flow methods \cite{flot,pointpwc,HPLFlowNet,pvraft,FlowStep3DCVPR2021}. We adopt \textit{EPE3D} (3D end-point-error) as  our primary evaluation metric that computes the average $\mathit{L}_2$ distance between the predicted and GT scene flow in meters, following existing methods such as  \cite{Mittal_2020_CVPR,weaklyrigidflow,randomwalk_sceneflow}. 

We also evaluate accuracy at three threshold levels: 
(1)	\textit{Acc3DS} (strict
accuracy) that computed the ratio of points whose EPE3D $<$ 0.05m or relative error $<$5\%.
(2) \textit{Acc3DR} (relaxed accuracy) which computes the ratio of points whose EPE3D $<$0.10m or relative error $<$10\%. (3) \textit{Outliers} that computes the ratio of points whose EPE3D $>$0.30m or relative error $>$10\%.

%%%%%%%%%%%%%%%%%%%%%%%%%%%%%%%%%%%%%%%%%%%%%%
\begin{figure*}[!htp]
\centering	
\centering	
\begin{minipage}[t]{0.21\textwidth}
\centering
\includegraphics[bb=0 0 1780 1360,width=1\textwidth]{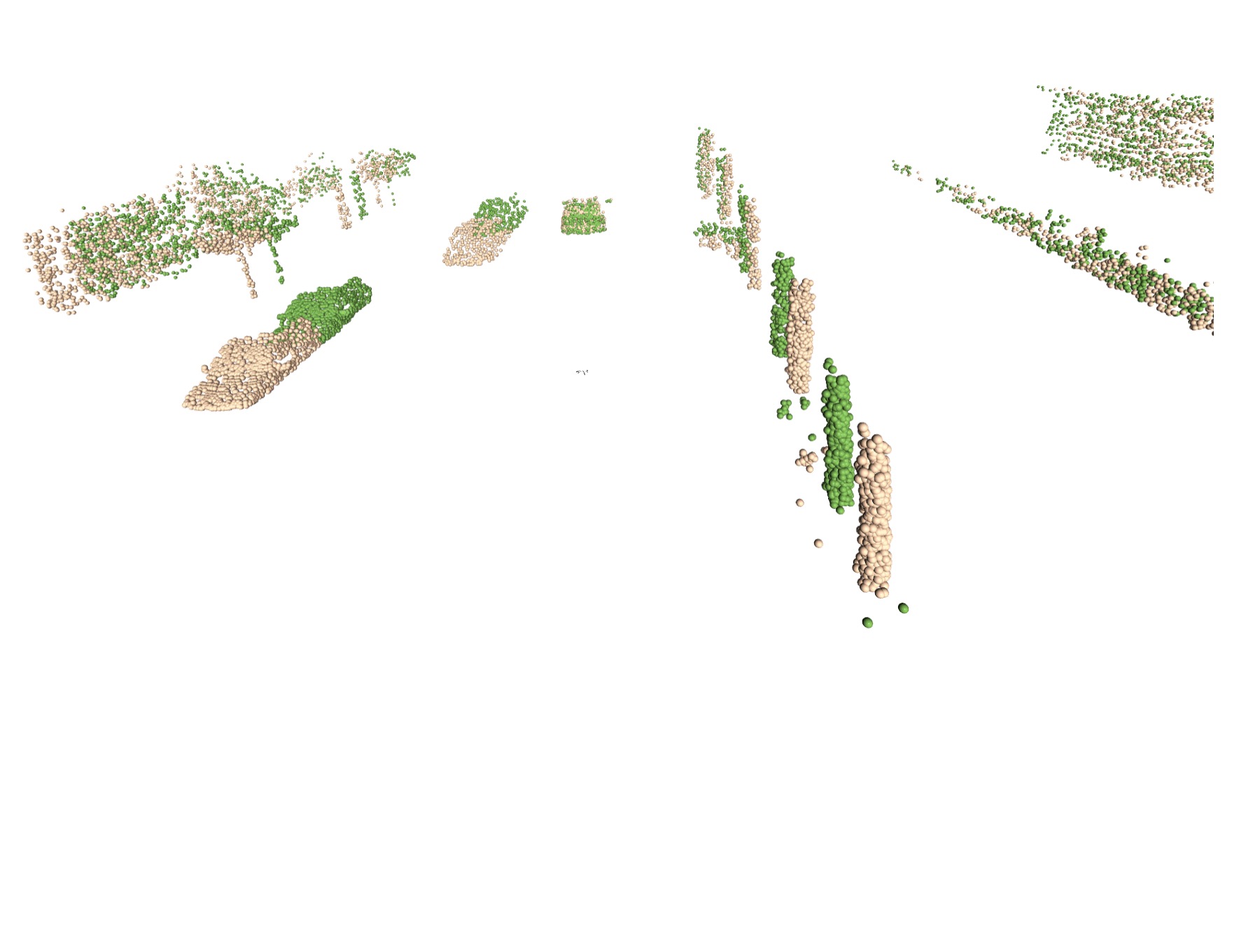}
\includegraphics[bb=0 0 1780 1360,width=1\textwidth]{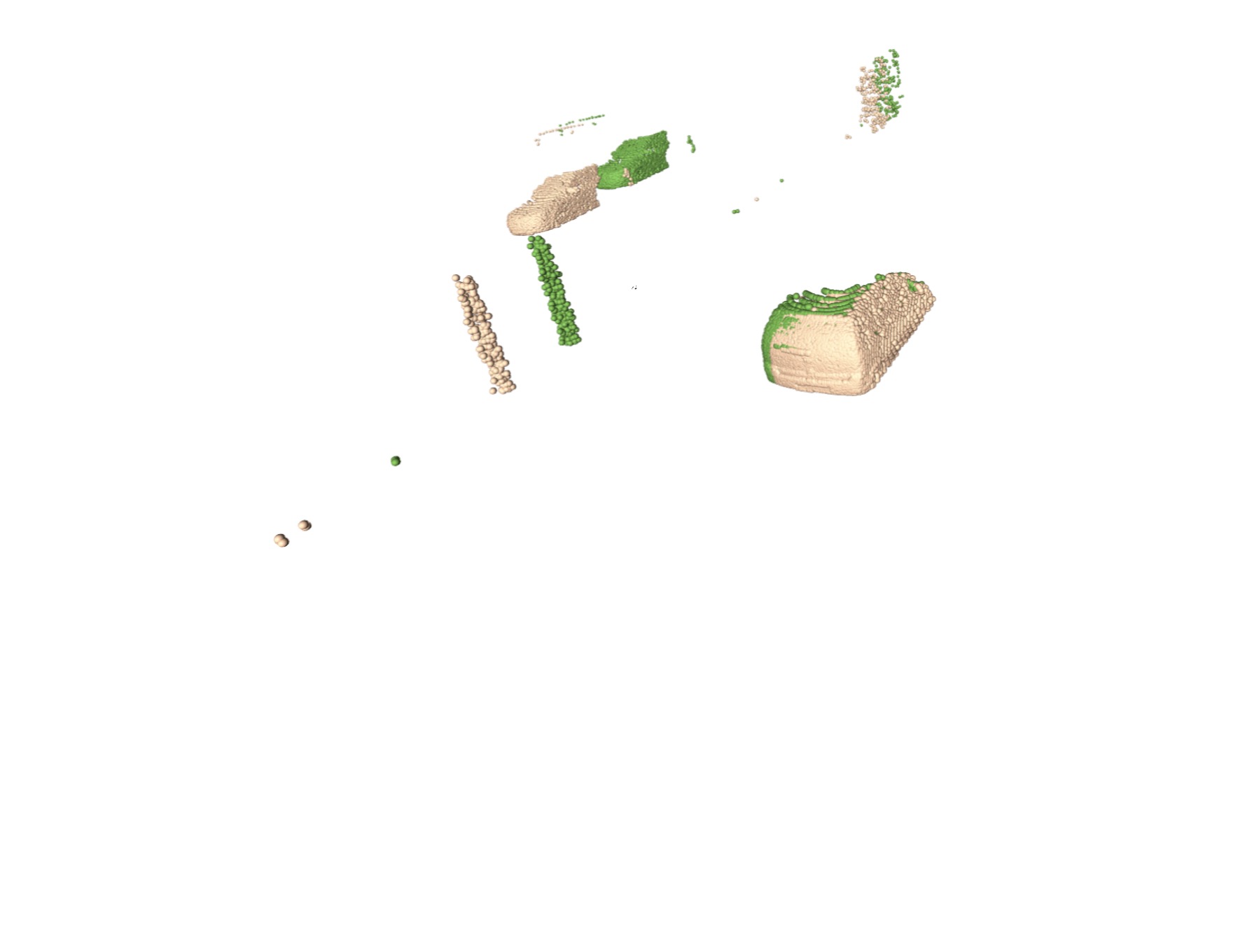}

\scriptsize (a) Original point clouds
\end{minipage}
\begin{minipage}[t]{0.21\textwidth}
\centering
\includegraphics[bb=0 0 1780 1360,width=1\textwidth]{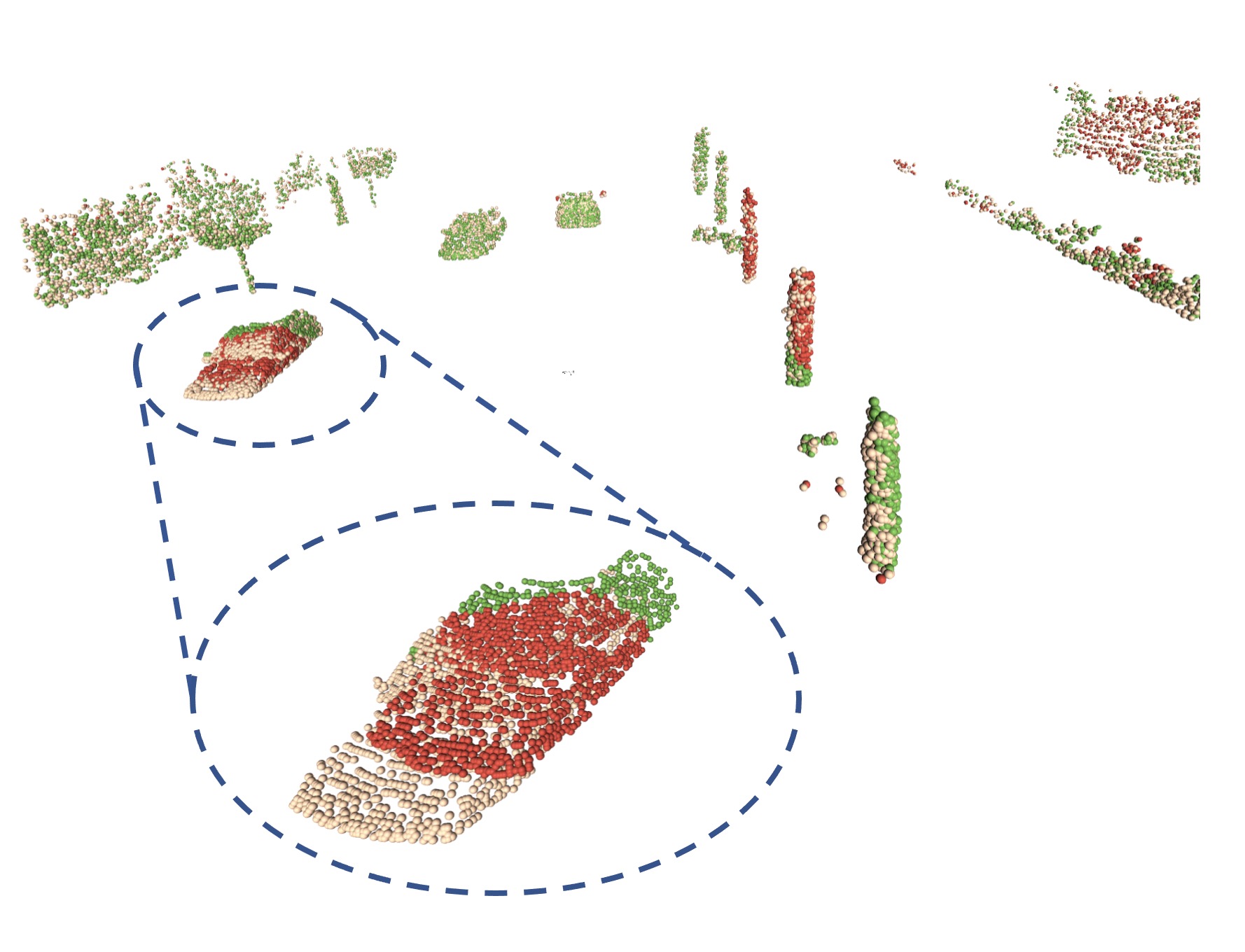}
\includegraphics[bb=0 0 1780 1360,width=1\textwidth]{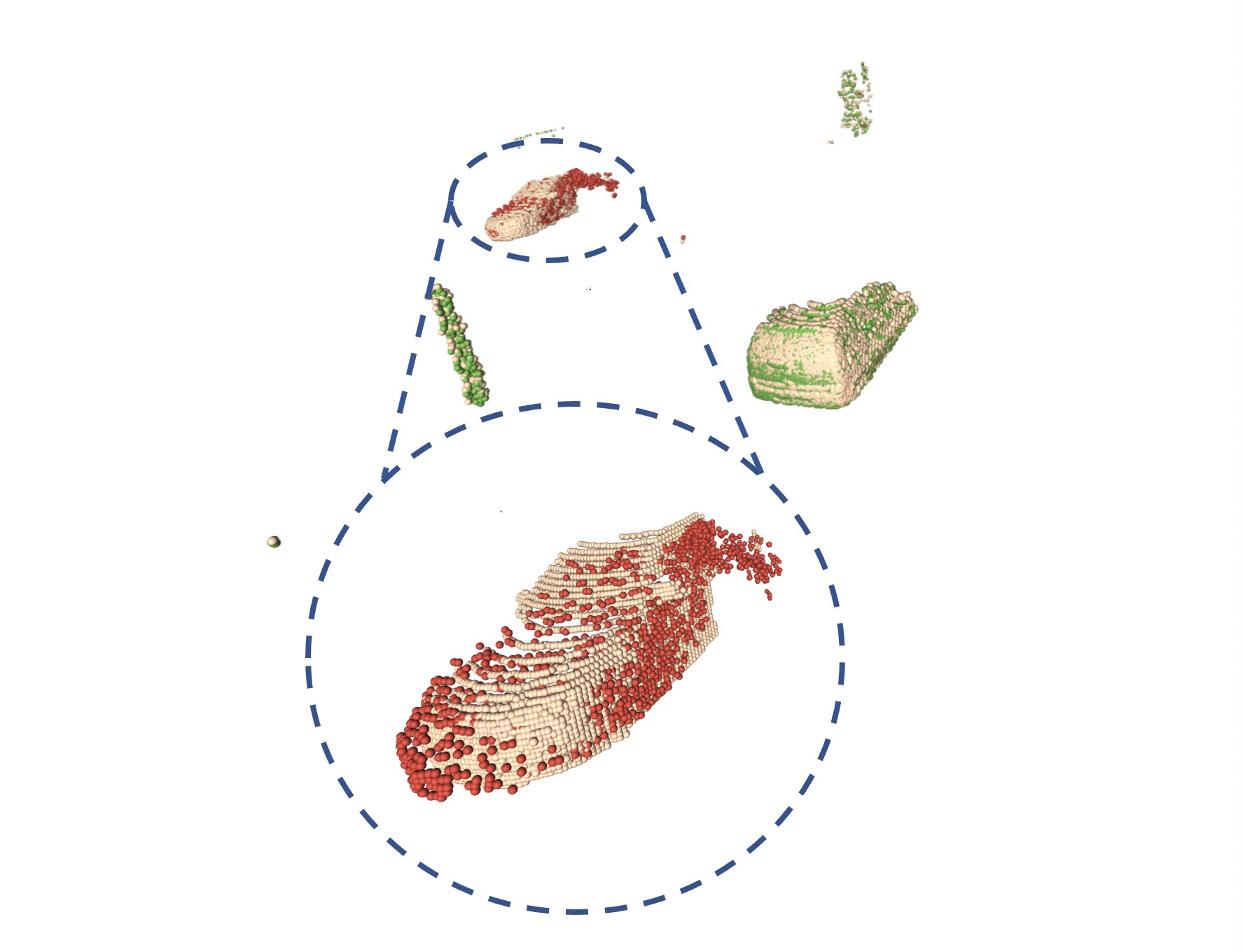}

\scriptsize (b) FlowStep3D \cite{FlowStep3DCVPR2021}
\end{minipage}
\begin{minipage}[t]{0.21\textwidth}
\centering
\includegraphics[bb=0 0 1780 1360,width=1\textwidth]{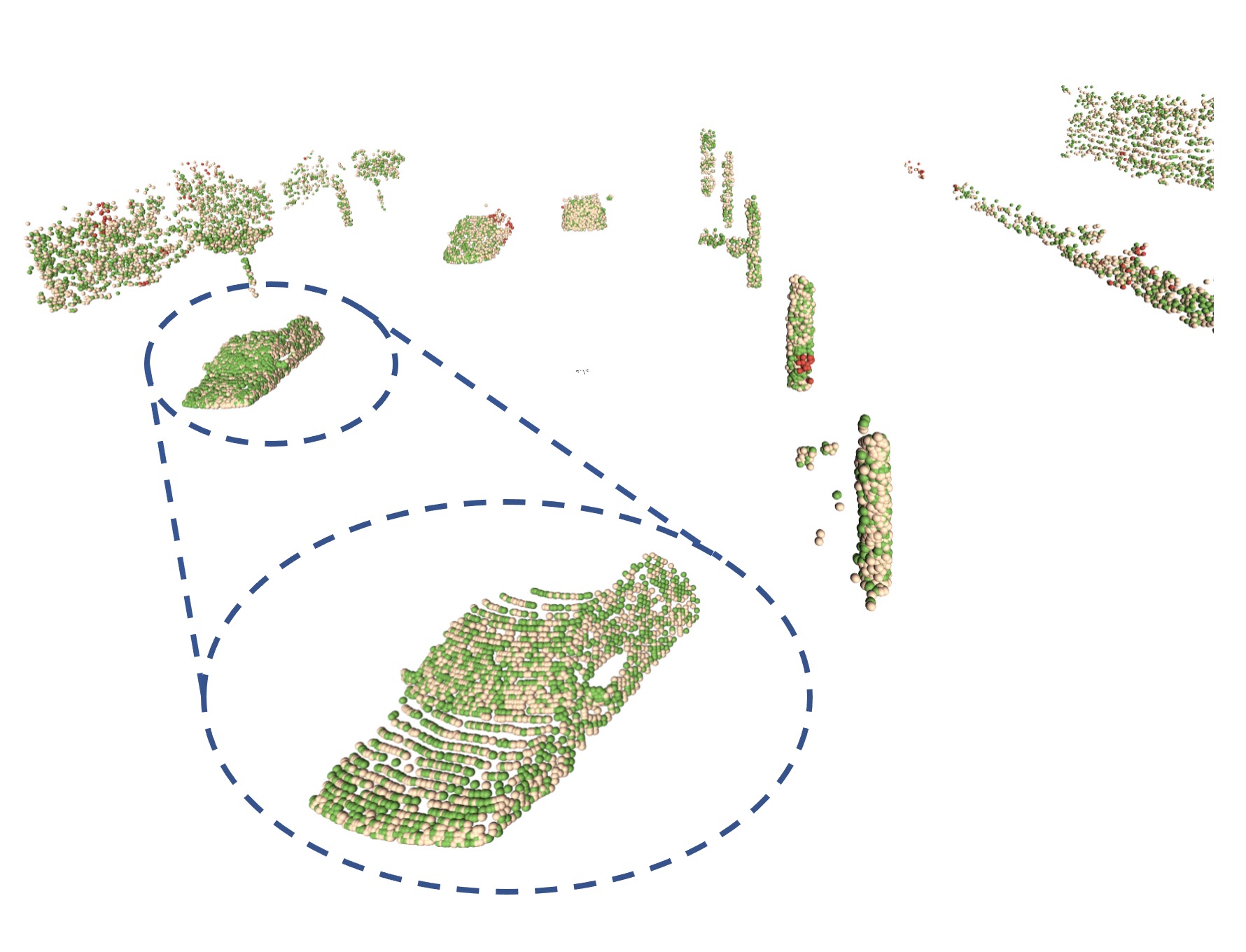}
\includegraphics[bb=0 0 1780 1360,width=1\textwidth]{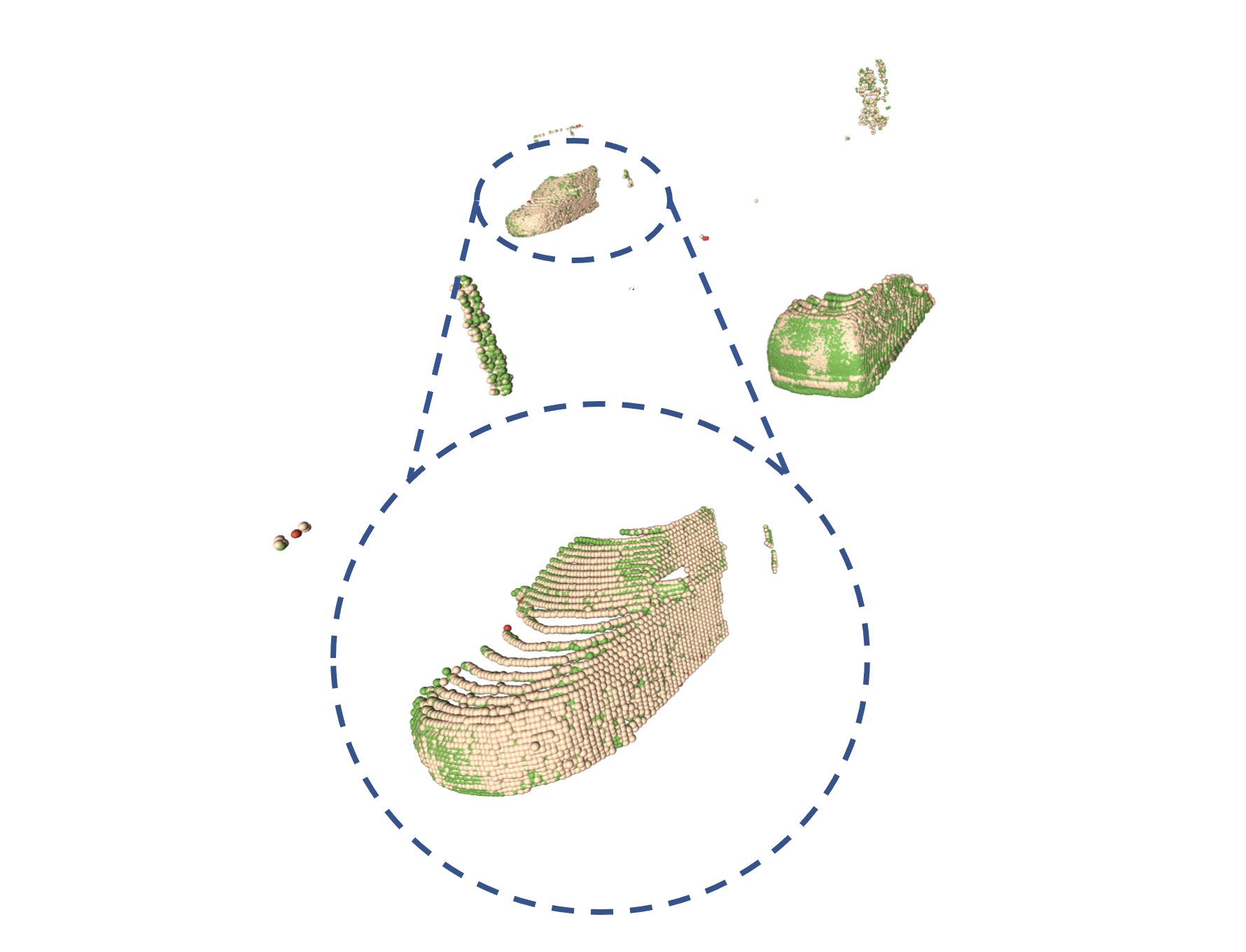}

\scriptsize (c) Ours

\end{minipage}
\begin{minipage}[t]{0.21\textwidth}
\centering
\includegraphics[bb=0 0 1780 1360,width=1\textwidth]{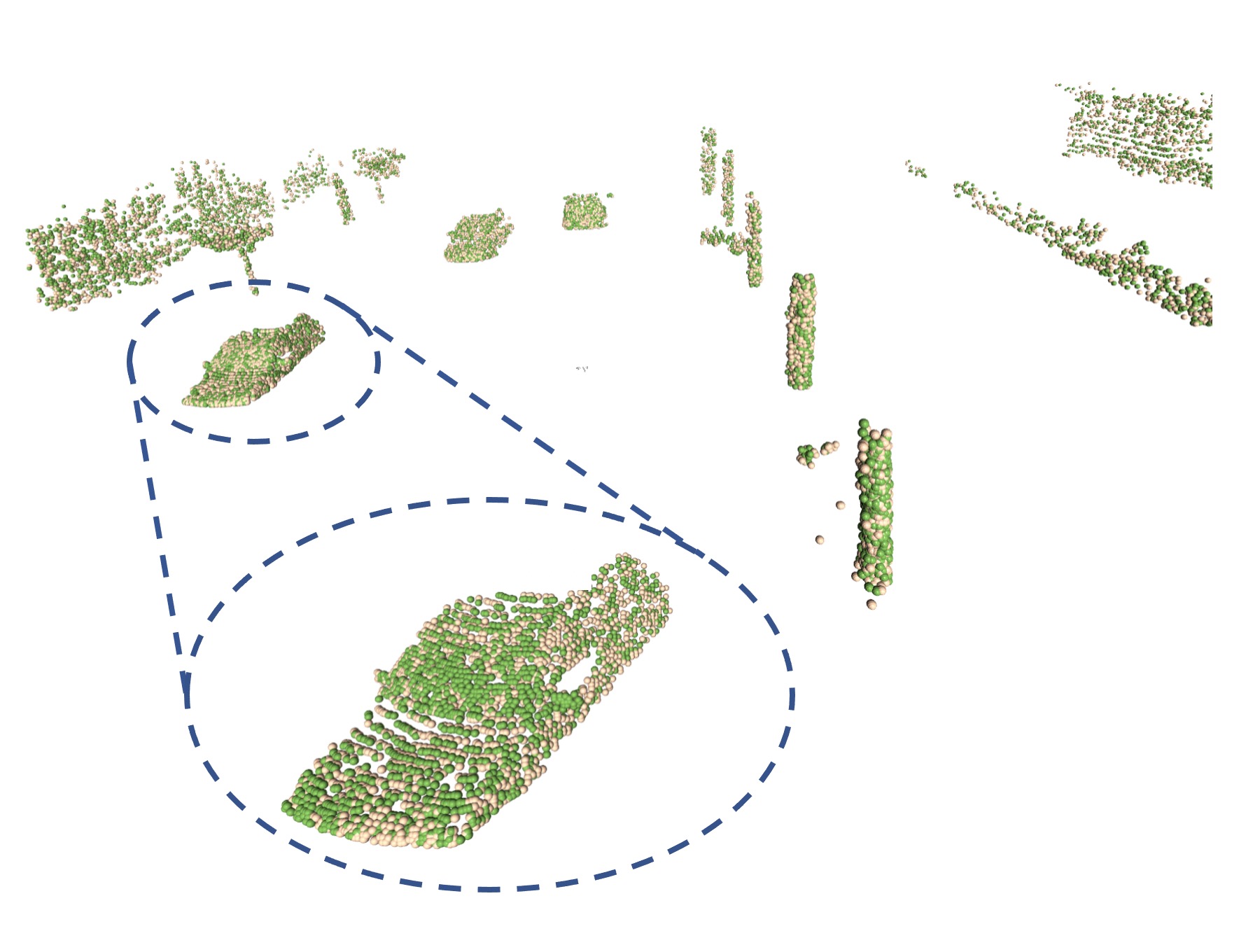}
\includegraphics[bb=0 0 1780 1360,width=1\textwidth]{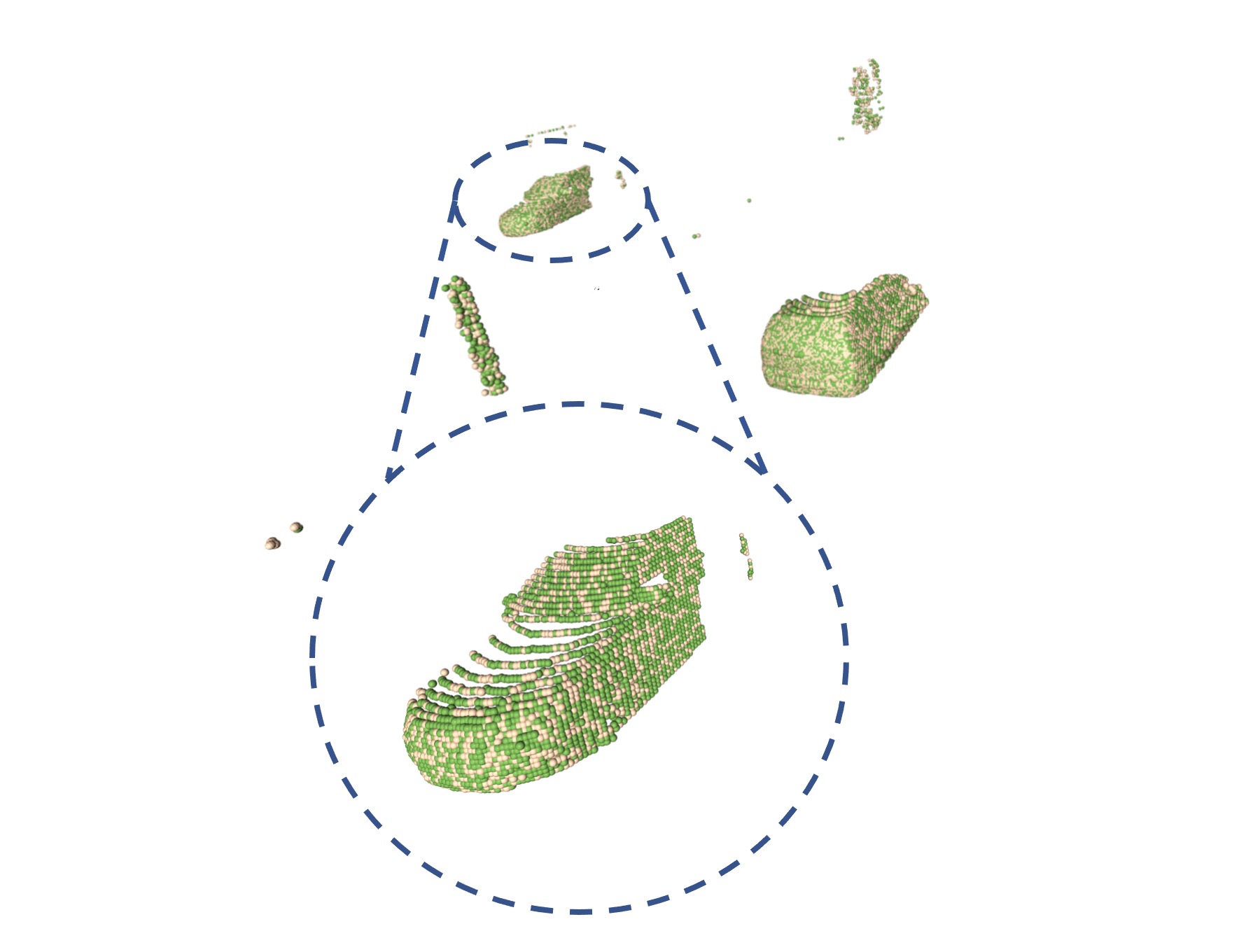}

\scriptsize (d) Ground-truth scene flow
\end{minipage}

\caption{\textbf{Qualitative comparison results on KITTI.} Orange points indicate the second point cloud.  Green points  indicate the first point cloud in (a), the first point cloud warped by correctly predicted flows in (b)(c), and the first point cloud warped by ground-truth flows in (d). Red points indicate the points warped by incorrectly predicted flows whose EPE3D $>$0.1m. Dotted circles zoom in some regions.
}

\label{fig: compared_self_supervised}
\vspace*{-0pt}%-12
\end{figure*}

%%%%%%%%%%%%%%%%%%%%%%%%%%%%%%%%%%%%%%%%%%%%%%
\begin{figure}[!htp]
\centering	
\begin{minipage}[t]{0.156\textwidth}
\centering
\includegraphics[bb=0 0 1638  918, trim=11cm 0.1cm 10cm 6cm,clip, width=1\textwidth]{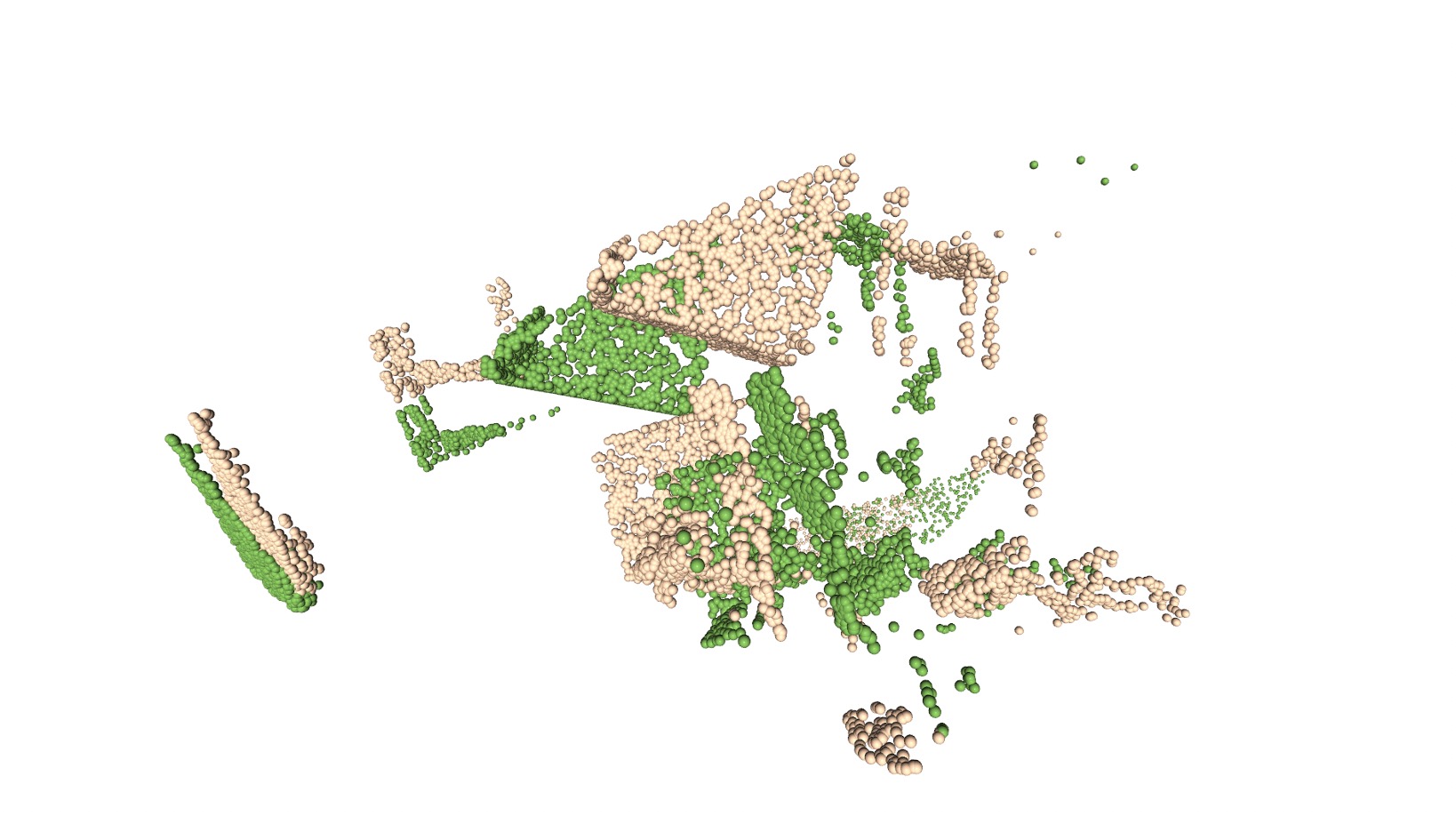}	
\scriptsize (a) Original point clouds
\end{minipage}
\begin{minipage}[t]{0.156\textwidth}
\centering
\includegraphics[bb=0 0 1638  918,trim=11cm 0.1cm 10cm 6cm,clip, width=1\textwidth]{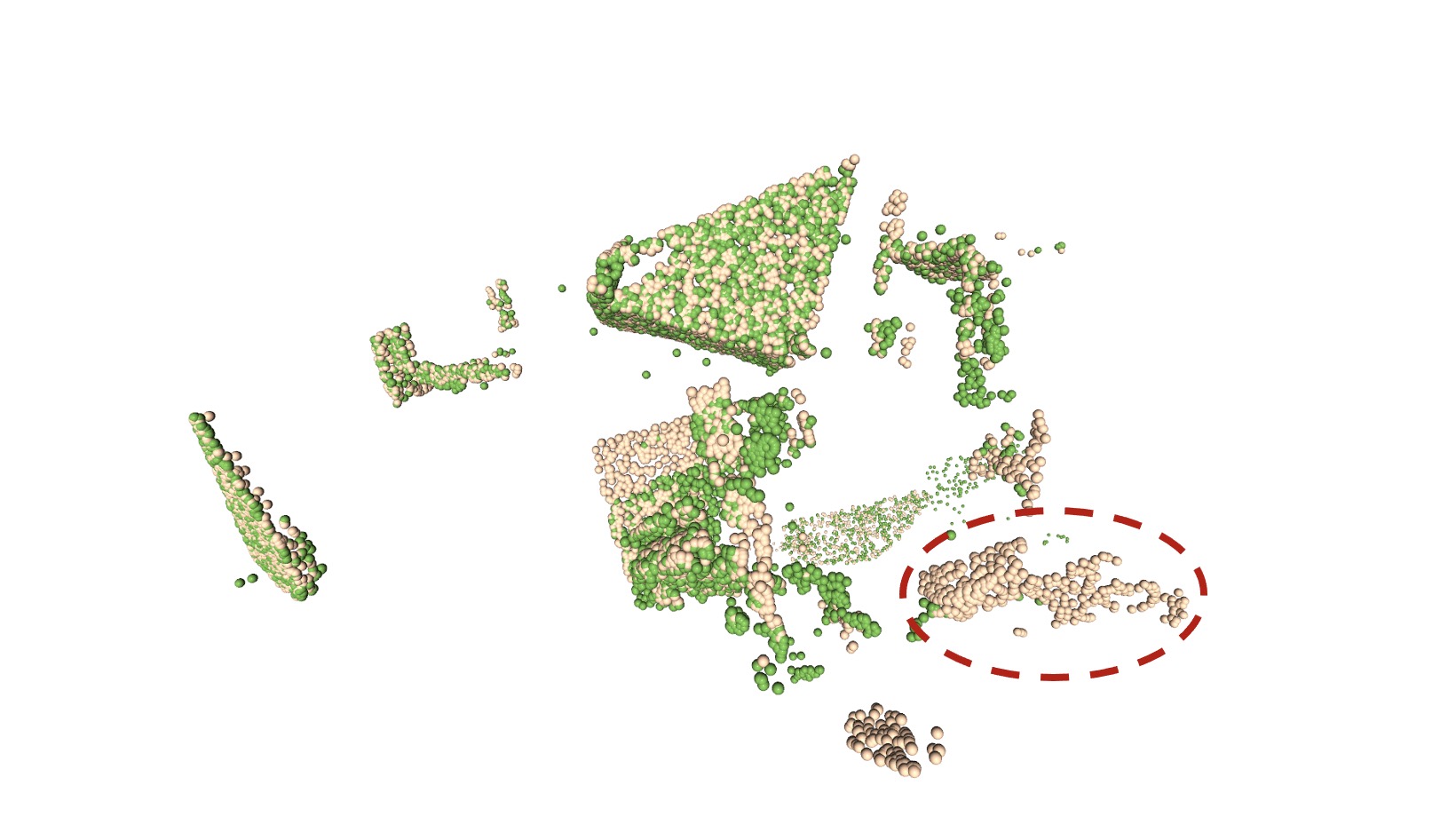}	
\scriptsize (b) Pseudo scene flow

\end{minipage}
\begin{minipage}[t]{0.156\textwidth}
\centering
\includegraphics[bb=0 0 1638  918, trim=11cm 0.1cm 10cm 6cm,clip, width=1\textwidth]{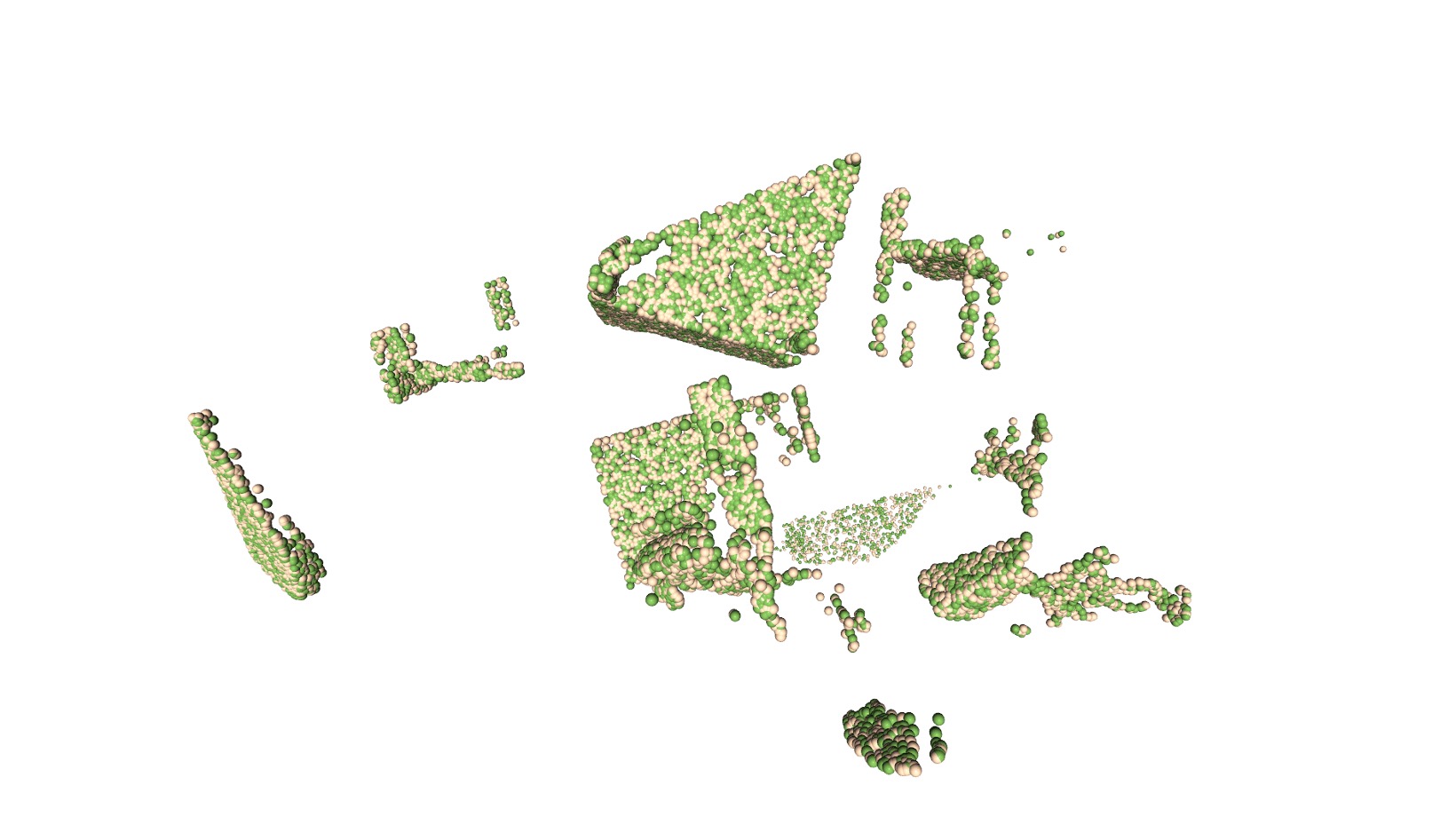}		
\scriptsize (c) Ground truth scene flow
\end{minipage}

\caption{\textbf{Visualize results  of pseudo scene flow on FT3D.} Orange points indicate the second point cloud. Green points indicate the first point cloud in (a), the first point cloud warped by pseudo scene flow in (b), and the first point cloud warped by ground truth flow in (c). 
} 

\label{fig: compared pseduo label}
\vspace*{-4pt}%-12
\end{figure}
%%%%%%%%%%%%%%%%%%%%%%%%%%%%%%%%%%%%%%%%%%%%%%

\mysection{Implementation Details}
Our scene flow network is based on state-of-the-art method FLOT \cite{flot}, where   the feature extractor is replaced by  Minkowski-Net \cite{minkowskiSPC}, due to its good feature representation in point cloud segmentation \cite{spvnas}\cite{PointContrast2020} and scene flow estimation \cite{weaklyrigidflow}. The voxel size is set to be 0.06m, and $\lambda=0.25$. The detailed architecture of our scene flow network is described in the supplementary.
Our method can be trained on any unlabeled datasets that contain synchronized monocular RGB images and point clouds. When generating pseudo scene flow labels for point clouds in FT3D dataset, we employ the optical flow estimation model \cite{teed2020raft} trained on FlyingChairs \cite{dosovitskiy2015flownet} for estimating optical flow, in order to circumvent any ground-truth annotations related to FT3D dataset.  
We implement our method in Pytorch \cite{paszke2019pytorch} and employ Adam
optimization algorithm for training. The learning rate begins with 0.001 for 50 epochs and is then linearly decayed.
Our model is trained on one NVIDIA V100 with batch size 5.

%%%%%%%%%%%%%%%%==========================
\begin{table}[t]
\centering
\small
\begin{tabular}{p{2.5cm} |p{1.4cm}<{\centering} | p{1.4cm}<{\centering} p{1.2cm}<{\centering} p{1.5cm}<{\centering}  }
\hline
Method &Sup. & EPE3D $\downarrow$ & AR $\uparrow$  \\
\hline
FLOT\cite{flot}  &    \textit{Full}    &   0.653&   0.313    \\
Rigid3DSF\cite{weaklyrigidflow} &    \textit{Full}    & 0.535 &  0.437   \\

Ours&       \textit{Self}                  &\bf0.165&\bf0.711 \\
\hline
\end{tabular}
\caption{\textbf{Comparison results  on a real-world LiDAR dataset L-KITTI.}  The Performance of 	Rigid3DSF\cite{weaklyrigidflow}  and  FLOT\cite{flot}  is
reported from \cite{weaklyrigidflow}.}
\label{tab:comparison_lidar}
\vspace*{-0pt}
\end{table} 
%%%%%%%%%%%%%%%%==========================

\subsection{Quantitative Evaluation}
Since our method does not use ground-truth scene flow, we mainly compare our method with state-of-the-art self-supervised methods including PointPWC\cite{pointpwc}, SelfPF \cite{randomwalk_sceneflow} and FlowStep3D \cite{FlowStep3DCVPR2021}. Tab. \ref{tab:comparison} also demonstrates the performance of recent representative supervised methods, which shows that our method even outperforms  some  state-of-art
supervised  approaches \eg \cite{flownet3d}\cite{HPLFlowNet}\cite{tishchenko2020self}.

\mysection{Results on FT3D}
The FT3D dataset contains monocular images and point clouds.
We use monocular images and point clouds from the training data of FT3D dataset to generate pseudo scene flow labels for the point clouds. With those pseudo labels, we train our scene flow network on point clouds. We then evaluate the trained scene flow network on the testing point clouds of FT3D, following the same setting of existing methods \cite{pointpwc,FlowStep3DCVPR2021}.

Tab. \ref{tab:comparison} shows our method outperforms all \textit{self-supervised} methods in all evaluation metrics on the FT3D dataset. Compared with most recent \textit{{self-supervised}} methods SelfPF and FlowStep3D, our method outperforms these methods on  EPE3D by 49\% and 32\%.
Moreover, the EPE3D value of our 
method is around 5cm, which is a large improvement over the self-supervised methods.

\mysection{Results on S-KITTI without Fine-tune}
We test our method on S-KITTI to evaluate the generalization ability of our method. 
Following existing methods \cite{pointpwc,FlowStep3DCVPR2021,flot}, we train our scene flow model on FT3D, and then directly evaluate the model on S-KITTI without fine-tuning. 
As shown in Tab. \ref{tab:comparison}, our method achieves the highest performance in all metrics, compared with all self-supervised method,   demonstrating that our method well generalizes to the S-KITTI dataset.
Our method outperforms these state-of-the-art methods by a large margin. For example, the EPE3D of  our method is below 0.06m.
Compared with  supervised methods, our method also achieves competitive performance, which are even on par with some supervised approaches \cite{flownet3d}\cite{flot}\cite{pvraft}\cite{tishchenko2020self}\cite{HPLFlowNet}.

\mysection{Results on real-world LiDAR dataset} L-KITTI is a real-world LiDAR dataset. We evaluate our method on L-KITTI to show the effectiveness of our method on real-world data. The ground-truth annotations of training data are unavailable on L-KITTI.
Thanks to the proposed pseudo label generation module, we can train our scene flow model on L-KITTI without ground-truth scene flows.
In contrast, since FLOT \cite{flot} is a supervised method and cannot be trained without ground truth. We test the performance of FLOT by using its model trained on FT3D.

Tab. \ref{tab:comparison_lidar} shows that our method achieves the best performance on  L-KITTI. Due to large domain gap between FT3D and L-KITTI, the performance of FLOT \cite{flot} is degraded.  
In contrast, our method allows us to train on L-KITTI, achieving better performance.
\subsection{Qualitative evaluation}
Fig. \ref{fig: compared_self_supervised} visualizes the scene flow estimation results of our method.
To qualitatively evaluate the quality of predicted scene flow, we warp the first point cloud $\mathcal{P}^{t}$ using the
predicted scene flow, and then visualize the warped point cloud. For better illustrating the results, we zoom in some regions and color the points that are warped by inaccurately predicted scene flows by red. 
A high-quality scene flow makes the warped point cloud well overlaps the second point cloud $\mathcal{P}^{t+1}$.
As shown in Fig.\ref{fig: compared_self_supervised}, our method predicts scene flow with higher quality, compared with state-of-the-art self-supervised method \cite{FlowStep3DCVPR2021}.

%%%%%%%%%%%%%%%%%%%%%%%%%%%===================

\begin{table}[t]
\centering
\small{
\begin{tabular}{p{1.0cm}| p{3.8 cm}|  p{1.2cm}<{\centering} p{0.7cm}<{\centering} }
\toprule
Dataset & Method & EPE3D $\downarrow$ & AR $\uparrow$  \\
\midrule
\multirow{3}{*}{FT3D} 
& Chamfer loss      &   0.150 &   0.539  \\
& Chamfer $+$ smoothness loss    &  0.120 &  0.636  \\
& Pseudo  label  based loss(Ours)  &\bf0.068 &\bf0.881 \\
\midrule
\multirow{3}{*}{S-KITTI} 
& Chamfer loss         &   0.131 &  0.746  \\
& Chamfer $+$ smoothness loss &   0.103 &  0.775  \\
& Pseudo  label  based loss(Ours)  &\bf0.058 &\bf0.898 \\
\bottomrule
\end{tabular}
}
\caption{\textbf{{Ablation Analysis} of multi-modality pseudo label generation on FT3D and S-KITTI.} Best results are in bold.}
\label{tab:self-supervised-loss}
\vspace*{-0pt}%-12
\end{table}

\subsection{Ablation studies}

\mysection{The effect of multi-modality pseudo label generation} 
To validate the effectiveness of our pseudo labels on training scene flow network, we compare with the learning manner in state-of-the-art self-supervised methods \cite{FlowStep3DCVPR2021,pointpwc}.
These methods \cite{FlowStep3DCVPR2021,pointpwc} use the chamfer loss as the proxy loss  which approximates pseudo flow labels only from point clouds.  
Following these methods, we build the first baseline that removes our pseudo scene flow labels from the training loss, and employs the chamfer loss as the main loss.  
We also build the second baseline that employs the  chamfer loss and the smoothness loss including smoothness constraint and Laplacian regularization like \cite{pointpwc}) (see details in \cite{pointpwc}).

Tab. \ref{tab:self-supervised-loss} shows that our method outperforms the two baselines by a significant margin. Compared with chamfer loss and smoothness loss, our pseudo scene flow generation leverages the information from both RGB images and point clouds, which generates more informative pseudo scene flow labels with higher quality. This enables our method to  better train the scene flow network.

\mysection{The effect of noisy-label-aware learning scheme}
We validate the effectiveness of noisy-label-aware learning scheme from two aspects.
First, the first baseline simply treats pseudo scene flow labels as predicted scene flow without training a scene flow network. EPE3D value in the first row of   Tab. \ref{tab:noise} shows that the accuracy of  pseudo scene labels is not high\footnote{The results of the first baseline are unavaible on S-KITTI, since we don't train our networks on this dataset and don't have its pseudo labels.}. Fig. \ref{fig: compared pseduo label} illustrates the visualization results of  pseudo scene flow labels. Pseudo scene flow labels are good on some regions, however, are still noisy, indicating the importance of noisy-label-aware learning scheme.

Second, we test the effectiveness of label noise detection module. We build a baseline by removing the noise detection module from our learning scheme. 
Tab. \ref{tab:noise} shows that this baseline achieves poor performance, since all pseudo labels are treated equally during training. As a result, inaccurate pseudo labels play an important role in training,   make the training unstable.
In contrast, our noise detection module detects inaccurate labels and assigns low confidences to these labels, which reduces the negative effect of inaccurate labels.

\begin{table}[t]
\centering
\small
\begin{tabular}{p{1.0cm}| p{1.0cm}<{\centering}  p{2.0cm}<{\centering}   |p{1.32cm}<{\centering} p{0.8cm}<{\centering} p{1.1cm}<{\centering}  }
\toprule
Dataset & Training & noise detection & EPE3D $\downarrow$ &AR $\uparrow$\\
%\toprule
%\bottomrule
\midrule
\multirow{3}{*}{FT3D} 
&      &   &   0.602  &0.714 \\
&\ding{51} &   &  0.131   & 0.635 \\
&          \ding{51}                &\ding{51}&\bf0.068 &\bf0.881\\
\midrule
\multirow{3}{*}{S-KITTI} 
&         &  & - &  - \\
&         \ding{51}              &   &  0.139  & 0.668\\
&             \ding{51}          &\ding{51} &\bf0.058 &\bf0.898 \\
\bottomrule
\end{tabular}
\caption{\textbf{Ablation study of the proposed training with noisy labels.} Best results are in bold.}
\label{tab:noise}
\vspace*{-5pt}
\end{table}

\section{Conclusions}
In this paper, we propose a novel scene flow estimation method to capture scene flow from point clouds, without reliance on GT scene flow labels. Our pseudo label generation module leverages monocular images to generate pseudo labels for point clouds, which facilitates the training of scene flow networks. Our training scheme effectively reduces the negative effect of noises in pseudo labels on the training. 
The experimental results demonstrate our method not only achieves the best performance in self-supervised approaches, but also outperforms some supervised methods.
The superior performance of our method on synthetic data and real-world LiDAR data highlights the effectiveness of our method. We show it is possible  to generate informative pseudo scene flow labels for point clouds using multi-sensor data, which can help scene flow estimation.

\clearpage
%%%%%%%%% REFERENCES
{\small
\bibliographystyle{ieee_fullname}
\bibliography{flow1}

\begin{thebibliography}{10}\itemsep=-1pt

\bibitem{black1993framework}
Michael~J Black and Padmanabhan Anandan.
\newblock A framework for the robust estimation of optical flow.
\newblock In {\em ICCV}, pages 231--236. IEEE, 1993.

\bibitem{brox2009large}
Thomas Brox, Christoph Bregler, and Jitendra Malik.
\newblock Large displacement optical flow.
\newblock In {\em CVPR}, pages 41--48. IEEE, 2009.

\bibitem{shapenet}
Angel~X. Chang, Thomas Funkhouser, Leonidas Guibas, Pat Hanrahan, Qixing Huang,
  Zimo Li, Silvio Savarese, Manolis Savva, Shuran Song, Hao Su, Jianxiong Xiao,
  Li Yi, and Fisher Yu.
\newblock {ShapeNet}: An information-rich {3D} model repository.
\newblock Technical Report arXiv:1512.03012 [cs.GR], arXiv preprint, 2015.

\bibitem{chen2020consistency}
Yuhua Chen, Luc Van~Gool, Cordelia Schmid, and Cristian Sminchisescu.
\newblock Consistency guided scene flow estimation.
\newblock In {\em ECCV}, pages 125--141. Springer, 2020.

\bibitem{minkowskiSPC}
Christopher Choy, JunYoung Gwak, and vio Savarese.
\newblock 4d spatio-temporal convnets: Minkowski convolutional neural networks.
\newblock In {\em CVPR}, pages 3075--3084, 2019.

\bibitem{dewan2016rigid}
Ayush Dewan, Tim Caselitz, Gian~Diego Tipaldi, and Wolfram Burgard.
\newblock Rigid scene flow for 3d lidar scans.
\newblock In {\em IROS}, pages 1765--1770. IEEE, 2016.

\bibitem{dosovitskiy2015flownet}
Alexey Dosovitskiy, Philipp Fischer, Eddy Ilg, Philip Hausser, Caner Hazirbas,
  Vladimir Golkov, Patrick Van Der~Smagt, Daniel Cremers, and Thomas Brox.
\newblock Flownet: Learning optical flow with convolutional networks.
\newblock In {\em ICCV}, pages 2758--2766, 2015.

\bibitem{geiger2013vision}
Andreas Geiger, Philip Lenz, Christoph Stiller, and Raquel Urtasun.
\newblock Vision meets robotics: The kitti dataset.
\newblock {\em The International Journal of Robotics Research},
  32(11):1231--1237, 2013.

\bibitem{geiger2012automatic}
Andreas Geiger, Frank Moosmann, {\"O}mer Car, and Bernhard Schuster.
\newblock Automatic camera and range sensor calibration using a single shot.
\newblock In {\em 2012 IEEE international conference on robotics and
  automation}, pages 3936--3943. IEEE, 2012.

\bibitem{weaklyrigidflow}
Zan Gojcic, Or Litany, Andreas Wieser, Leonidas~J. Guibas, and Tolga Birdal.
\newblock Weakly supervised learning of rigid 3d scene flow.
\newblock In {\em CVPR}, pages 5692--5703, June 2021.

\bibitem{HPLFlowNet}
Xiuye Gu, Yijie Wang, Chongruo Wu, Yong~Jae Lee, and Panqu Wang.
\newblock Hplflownet: Hierarchical permutohedral lattice flownet for scene flow
  estimation on large-scale point clouds.
\newblock In {\em CVPR}, 2019.

\bibitem{horn1981determining}
Berthold~KP Horn and Brian~G Schunck.
\newblock Determining optical flow.
\newblock {\em Artificial intelligence}, 17(1-3):185--203, 1981.

\bibitem{huguet2007variational}
Fr{\'e}d{\'e}ric Huguet and Fr{\'e}d{\'e}ric Devernay.
\newblock A variational method for scene flow estimation from stereo sequences.
\newblock In {\em ICCV}, pages 1--7. IEEE, 2007.

\bibitem{hui2020liteflownet3}
Tak-Wai Hui and Chen~Change Loy.
\newblock Liteflownet3: Resolving correspondence ambiguity for more accurate
  optical flow estimation.
\newblock In {\em ECCV}, pages 169--184. Springer, 2020.

\bibitem{hui2018liteflownet}
Tak-Wai Hui, Xiaoou Tang, and Chen~Change Loy.
\newblock Liteflownet: A lightweight convolutional neural network for optical
  flow estimation.
\newblock In {\em CVPR}, pages 8981--8989, 2018.

\bibitem{Hur:2020:SSM}
Junhwa Hur and Stefan Roth.
\newblock Self-supervised monocular scene flow estimation.
\newblock In {\em CVPR}, 2020.

\bibitem{Hur:2021:SSM}
Junhwa Hur and Stefan Roth.
\newblock Self-supervised multi-frame monocular scene flow.
\newblock In {\em CVPR}, 2021.

\bibitem{ilg2017flownet}
Eddy Ilg, Nikolaus Mayer, Tonmoy Saikia, Margret Keuper, Alexey Dosovitskiy,
  and Thomas Brox.
\newblock Flownet 2.0: Evolution of optical flow estimation with deep networks.
\newblock In {\em CVPR}, pages 2462--2470, 2017.

\bibitem{ilg2018occlusions}
Eddy Ilg, Tonmoy Saikia, Margret Keuper, and Thomas Brox.
\newblock Occlusions, motion and depth boundaries with a generic network for
  disparity, optical flow or scene flow estimation.
\newblock In {\em ECCV}, pages 614--630, 2018.

\bibitem{jiang2019sense}
Huaizu Jiang, Deqing Sun, Varun Jampani, Zhaoyang Lv, Erik Learned-Miller, and
  Jan Kautz.
\newblock Sense: A shared encoder network for scene-flow estimation.
\newblock In {\em ICCV}, pages 3195--3204, 2019.

\bibitem{FlowStep3DCVPR2021}
Yair Kittenplon, Yonina~C. Eldar, and Dan Raviv.
\newblock Flowstep3d: Model unrolling for self-supervised scene flow
  estimation.
\newblock In {\em CVPR}, pages 4114--4123, 2021.

\bibitem{FuseSeg}
Georg Krispel, Michael Opitz, Georg Waltner, Horst Possegger, and Horst
  Bischof.
\newblock Fuseseg: Lidar point cloud segmentation fusing multi-modal data.
\newblock In {\em Proc. of the IEEE Winter Conference on Applications of
  Computer Vision (WACV)}, 2020.

\bibitem{Li2021CVPR12}
Ruibo Li, Guosheng Lin, Tong He, Fayao Liu, and Chunhua Shen.
\newblock {HCRF-Flow}: Scene flow from point clouds with continuous high-order
  {CRFs} and position-aware flow embedding.
\newblock In {\em IEEE Conference on Computer Vision and Pattern Recognition
  (CVPR'21)}, 2021.

\bibitem{randomwalk_sceneflow}
Ruibo Li, Guosheng Lin, and Lihua Xie.
\newblock Self-point-flow: Self-supervised scene flow estimation from point
  clouds with optimal transport and random walk.
\newblock In {\em Proceedings of the IEEE/CVF Conference on Computer Vision and
  Pattern Recognition (CVPR)}, pages 15577--15586, June 2021.

\bibitem{Liu:2019:unrigid}
Wenlong Ye Yong~Liu Liang~Liu, Guangyao~Zhai.
\newblock Unsupervised learning of scene flow estimation fusing with local
  rigidity.
\newblock In {\em International Joint Conference on Artificial Intelligence,
  IJCAI}, 2019.

\bibitem{flownet3d}
Xingyu Liu, Charles~R Qi, and Leonidas~J Guibas.
\newblock Flownet3d: Learning scene flow in 3d point clouds.
\newblock In {\em CVPR}, 2019.

\bibitem{ma2019drisf}
Wei-Chiu Ma, Shenlong Wang, Rui Hu, Yuwen Xiong, and Raquel Urtasun.
\newblock Deep rigid instance scene flow.
\newblock In {\em CVPR}, 2019.

\bibitem{mayer2016large}
Nikolaus Mayer, Eddy Ilg, Philip Hausser, Philipp Fischer, Daniel Cremers,
  Alexey Dosovitskiy, and Thomas Brox.
\newblock A large dataset to train convolutional networks for disparity,
  optical flow, and scene flow estimation.
\newblock In {\em CVPR}, pages 4040--4048, 2016.

\bibitem{Menze2018JPRS}
Moritz Menze, Christian Heipke, and Andreas Geiger.
\newblock Object scene flow.
\newblock {\em JPRS}, 2018.

\bibitem{meyer2019sensor}
Gregory~P Meyer, Jake Charland, Darshan Hegde, Ankit Laddha, and Carlos
  Vallespi-Gonzalez.
\newblock Sensor fusion for joint 3d object detection and semantic
  segmentation.
\newblock In {\em Proceedings of the IEEE/CVF Conference on Computer Vision and
  Pattern Recognition Workshops}, pages 0--0, 2019.

\bibitem{Mittal_2020_CVPR}
Himangi Mittal, Brian Okorn, and David Held.
\newblock Just go with the flow: Self-supervised scene flow estimation.
\newblock In {\em Proceedings of the IEEE/CVF Conference on Computer Vision and
  Pattern Recognition (CVPR)}, June 2020.

\bibitem{MIFDB16}
N.Mayer, E.Ilg, P.H{\"a}usser, P.Fischer, D.Cremers, A.Dosovitskiy, and T.Brox.
\newblock A large dataset to train convolutional networks for disparity,
  optical flow, and scene flow estimation.
\newblock In {\em CVPR}, 2016.

\bibitem{flythings3d}
N.Mayer, E.Ilg, P.H{\"a}usser, P.Fischer, D.Cremers, A.Dosovitskiy, and T.Brox.
\newblock A large dataset to train convolutional networks for disparity,
  optical flow, and scene flow estimation.
\newblock In {\em CVPR}, 2016.

\bibitem{paszke2019pytorch}
Adam Paszke, Sam Gross, Francisco Massa, Adam Lerer, James Bradbury, Gregory
  Chanan, Trevor Killeen, Zeming Lin, Natalia Gimelshein, Luca Antiga, et~al.
\newblock Pytorch: An imperative style, high-performance deep learning library.
\newblock {\em arXiv preprint arXiv:1912.01703}, 2019.

\bibitem{flot}
Gilles Puy, Alexandre Boulch, and Renaud Marlet.
\newblock {FLOT}: {S}cene {F}low on {P}oint {C}louds {G}uided by {O}ptimal
  {T}ransport.
\newblock In {\em ECCV}, 2020.

\bibitem{qi2020imvotenet}
Charles~R Qi, Xinlei Chen, Or Litany, and Leonidas~J Guibas.
\newblock Imvotenet: Boosting 3d object detection in point clouds with image
  votes.
\newblock In {\em IEEE Conference on Computer Vision and Pattern Recognition
  (CVPR)}, 2020.

\bibitem{quiroga2014dense}
Julian Quiroga, Thomas Brox, Fr{\'e}d{\'e}ric Devernay, and James Crowley.
\newblock Dense semi-rigid scene flow estimation from rgbd images.
\newblock In {\em ECCV}, pages 567--582. Springer, 2014.

\bibitem{ranftl2014non}
Ren{\'e} Ranftl, Kristian Bredies, and Thomas Pock.
\newblock Non-local total generalized variation for optical flow estimation.
\newblock In {\em ECCV}, pages 439--454. Springer, 2014.

\bibitem{sun2015layered}
Deqing Sun, Erik~B Sudderth, and Hanspeter Pfister.
\newblock Layered rgbd scene flow estimation.
\newblock In {\em CVPR}, pages 548--556, 2015.

\bibitem{PWCNet}
Deqing Sun, Xiaodong Yang, Ming-Yu Liu, and Jan Kautz.
\newblock Pwc-net: Cnns for optical flow using pyramid, warping, and cost
  volume.
\newblock In {\em CVPR}, 2018.

\bibitem{spvnas}
Haotian* Tang, Zhijian* Liu, Shengyu Zhao, Yujun Lin, Ji Lin, Hanrui Wang, and
  Song Han.
\newblock Searching efficient 3d architectures with sparse point-voxel
  convolution.
\newblock In {\em ECCV}, 2020.

\bibitem{teed2020raft-3d}
Zachary Teed and Jia Deng.
\newblock Raft-3d: Scene flow using rigid-motion embeddings.
\newblock {\em arXiv preprint arXiv:2012.00726}, 2020.

\bibitem{teed2020raft}
Zachary Teed and Jia Deng.
\newblock Raft: Recurrent all-pairs field transforms for optical flow.
\newblock In {\em ECCV}, pages 402--419. Springer, 2020.

\bibitem{tishchenko2020self}
Ivan Tishchenko, Sandro Lombardi, Martin~R Oswald, and Marc Pollefeys.
\newblock Self-supervised learning of non-rigid residual flow and ego-motion.
\newblock {\em arXiv preprint arXiv:2009.10467}, 2020.

\bibitem{ushani2017learning}
Arash~K Ushani, Ryan~W Wolcott, Jeffrey~M Walls, and Ryan~M Eustice.
\newblock A learning approach for real-time temporal scene flow estimation from
  lidar data.
\newblock In {\em ICRA}, pages 5666--5673. IEEE, 2017.

\bibitem{vogel2013piecewise}
Christoph Vogel, Konrad Schindler, and Stefan Roth.
\newblock Piecewise rigid scene flow.
\newblock In {\em ICCV}, pages 1377--1384, 2013.

\bibitem{FESTA}
Haiyan Wang, Jiahao Pang, Muhammad~A. Lodhi, Yingli Tian, and Dong Tian.
\newblock Festa: Flow estimation via spatial-temporal attention for scene point
  clouds.
\newblock In {\em Proceedings of the IEEE/CVF Conference on Computer Vision and
  Pattern Recognition (CVPR)}, pages 14173--14182, June 2021.

\bibitem{FlowNet3Dplus}
Zirui Wang, Shuda Li, Henry Howard-Jenkins, Victor Prisacariu, and Min Chen.
\newblock Flownet3d++: Geometric losses for deep scene flow estimation.
\newblock In {\em WACV}, March 2020.

\bibitem{wedel2008efficient}
Andreas Wedel, Clemens Rabe, Tobi Vaudrey, Thomas Brox, Uwe Franke, and Daniel
  Cremers.
\newblock Efficient dense scene flow from sparse or dense stereo data.
\newblock In {\em ECCV}, pages 739--751. Springer, 2008.

\bibitem{pvraft}
Yi Wei, Ziyi Wang, Yongming Rao, Jiwen Lu, and Jie Zhou.
\newblock {PV-RAFT: Point-Voxel Correlation Fields for Scene Flow Estimation of
  Point Clouds}.
\newblock In {\em CVPR}, 2021.

\bibitem{weinzaepfel2013deepflow}
Philippe Weinzaepfel, Jerome Revaud, Zaid Harchaoui, and Cordelia Schmid.
\newblock Deepflow: Large displacement optical flow with deep matching.
\newblock In {\em ICCV}, pages 1385--1392, 2013.

\bibitem{pointpwc}
Wenxuan Wu, Zhi~Yuan Wang, Zhuwen Li, Wei Liu, and Li Fuxin.
\newblock Pointpwc-net: Cost volume on point clouds for (self-) supervised
  scene flow estimation.
\newblock In {\em ECCV}, pages 88--107, 2020.

\bibitem{PointContrast2020}
Saining Xie, Jiatao Gu, Demi Guo, Charles~R. Qi, Leonidas Guibas, and Or
  Litany.
\newblock Pointcontrast: Unsupervised pre-training for 3d point cloud
  understanding.
\newblock In {\em ECCV}, 2020.

\bibitem{zach2007duality}
Christopher Zach, Thomas Pock, and Horst Bischof.
\newblock A duality based approach for realtime tv-l 1 optical flow.
\newblock In {\em Joint pattern recognition symposium}, pages 214--223.
  Springer, 2007.

\bibitem{mmMOT2019ICCV}
Wenwei Zhang, Hui Zhou, Shuyang Sun, Zhe Wang, Jianping Shi, and Chen~Change
  Loy.
\newblock Robust multi-modality multi-object tracking.
\newblock In {\em The IEEE International Conference on Computer Vision (ICCV)},
  October 2019.

\bibitem{Zhuang_2021_ICCV}
Zhuangwei Zhuang, Rong Li, Kui Jia, Qicheng Wang, Yuanqing Li, and Mingkui Tan.
\newblock Perception-aware multi-sensor fusion for 3d lidar semantic
  segmentation.
\newblock In {\em Proceedings of the IEEE/CVF International Conference on
  Computer Vision (ICCV)}, pages 16280--16290, October 2021.

\end{thebibliography}
}

\end{document}